\documentclass{article}
\usepackage{iclr2027_conference,times}

\iclrfinalcopy   %

\usepackage[T1]{fontenc}
\usepackage{graphicx}
\usepackage{booktabs}
\usepackage{tabularx}
\usepackage{array}
\usepackage{multirow}
\usepackage{makecell}
\usepackage{amsmath}
\usepackage{amssymb}
\usepackage{bm}
\usepackage{xcolor}
\usepackage{colortbl}
\usepackage{enumitem}
\usepackage{listings}
\usepackage{pifont}
\usepackage{nicefrac}
\usepackage{subcaption}
\usepackage{capt-of}   %
\usepackage{xspace}
\usepackage{placeins}
\usepackage{titletoc}  %
\usepackage{wrapfig}
\usepackage{microtype}
\usepackage{hyperref}
\usepackage{url}
\usepackage{xurl}      %
\usepackage[scaled=0.92]{helvet} %
\usepackage[most]{tcolorbox}
\usepackage{fontawesome5}

\usepackage[nointegrals]{wasysym}   %
\definecolor{okgreen}{RGB}{34,139,34}
\definecolor{nored}{RGB}{200,40,40}
\definecolor{partial}{RGB}{214,140,0}
\newcommand{\cmark}{\textcolor{okgreen}{\ding{51}}}
\newcommand{\xmark}{\textcolor{nored}{\ding{55}}}
\newcommand{\pmark}{\textcolor{partial}{\LEFTcircle}}

\newcommand{\na}{\textcolor{gray}{---}}
\newcommand{\unit}[1]{\textcolor{gray}{#1}}
\definecolor{cvBlue}{HTML}{2F6FAD}
\definecolor{cvOrange}{HTML}{DD7B33}
\definecolor{cvGreen}{HTML}{3F8F5B}
\definecolor{cvRed}{HTML}{B23B3B}
\definecolor{cvPurple}{HTML}{7A5AA6}
\definecolor{cvGold}{HTML}{C08A1E}
\definecolor{cvTeal}{HTML}{2E8C99}
\definecolor{cvGrey}{HTML}{8C8C8C}
\definecolor{cvInk}{HTML}{2B2B2B}

\newcommand{\ourmethod}{3DCodeVerse}
\newcommand{\corpus}{\mbox{3DCodeVerse}-2M\xspace}
\newcommand{\harness}{\mbox{3DCodeVerse} Harness\xspace}
\newcommand{\bench}{\mbox{3DCodeBench}\xspace}
\newcolumntype{Y}{>{\raggedright\arraybackslash}X}
\newcolumntype{L}[1]{>{\raggedright\arraybackslash}p{#1}}
\newcolumntype{Z}[1]{>{\raggedright\arraybackslash\hsize=#1\hsize}X}

\lstdefinestyle{plain}{
  basicstyle=\ttfamily\scriptsize, breaklines=true, columns=fullflexible,
  keepspaces=true, showstringspaces=false, frame=single,
  rulecolor=\color[rgb]{0.75,0.75,0.75}, framesep=5pt, xleftmargin=6pt,
  commentstyle=\color[rgb]{0.35,0.35,0.35},
  escapeinside={(*@}{@*)},
}
\hypersetup{colorlinks=true, linkcolor=black, citecolor=black,
            urlcolor=[rgb]{0.10,0.25,0.65}}
\renewcommand{\headrulewidth}{0.4pt}
\definecolor{ypBlue}{HTML}{4A90D9}
\definecolor{gptastraPurple}{HTML}{8A5CB5}
\newcommand{\todo}[1]{\textcolor{red}{\textbf{[TODO: #1]}}}

\definecolor{cardBg}{HTML}{F0F6FD}
\definecolor{cardLink}{HTML}{D6337F}
\newcommand{\afl}[1]{\textsuperscript{#1}}
\newcommand{\cardau}[2]{\mbox{#1\afl{#2}}}  %
\newcommand{\cardicon}[1]{\makebox[1.5em][c]{\raisebox{0.65pt}{$\vcenter{\hbox{#1}}$}}\hspace{0.3em}}
\newcommand{\cardlink}[3]{\cardicon{#1}\textbf{\sffamily #2:}~\href{#3}{\texttt{#3}}}
\newcommand{\cvicon}{\includegraphics[height=0.95em]{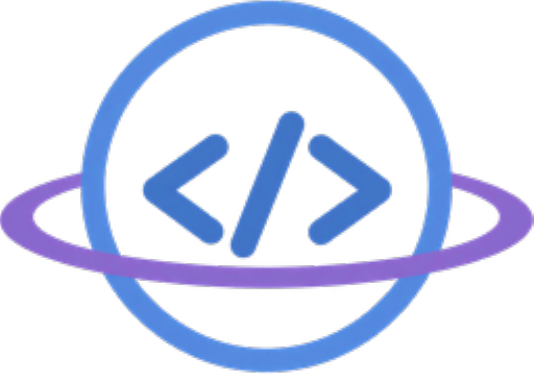}}  %
\newcommand{\hficon}{\includegraphics[height=0.98em]{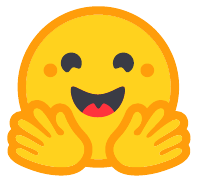}}

\begin{document}

\title{\ourmethod: Exploring 3D Worlds as Code}

  \begin{tcolorbox}[enhanced, colback=cardBg, colframe=cardBg, arc=4mm, boxrule=0pt,
    left=6mm, right=6mm, top=5mm, bottom=4mm, width=\textwidth]
  {\raggedright\sffamily\bfseries\fontsize{17}{21}\selectfont \ourmethod: Exploring 3D Worlds as Code\par}
  \vspace{3mm}
  {\raggedright\sffamily\bfseries\small
  \cardau{Yipeng Gao}{1*\dag}, \cardau{Ziyao Zeng}{2*}, \cardau{Jinfa Huang}{3*}, \cardau{Yijiang Li}{4*},
  \cardau{Haolin Xiong}{1}, \cardau{Wang Qin}{1}, \cardau{Xinlei Yu}{1}, \cardau{Youheng Yao}{2},
  \cardau{Jie Zhang}{5}, \cardau{Kun-Yu Lin}{6}, \cardau{Rong Liu}{1}, \cardau{Yinyuan Zhao}{7},
  \cardau{Yajie Zhao}{1}, \cardau{Meiqi Guo}{8}, \cardau{Jiebo Luo}{3}, \cardau{Lei Shu}{8},
  \cardau{Yunhao Ge}{1}, \cardau{Xinyu Hu}{9}, \cardau{Laurent Itti}{1}\par}
  \vspace{3mm}
  {\raggedright\small
  \mbox{\afl{1}University of Southern California}, \mbox{\afl{2}Yale University},
  \mbox{\afl{3}University of Rochester}, \mbox{\afl{4}University of California, San Diego},
  \mbox{\afl{5}Columbia University}, \mbox{\afl{6}Nanyang Technological University},
  \mbox{\afl{7}University of California, Santa Cruz}, \mbox{\afl{8}Google DeepMind},
  \mbox{\afl{9}Philo Labs}\par
  \vspace{0.8mm}
  {\footnotesize\afl{*}Equal contribution,\enspace \afl{\dag}Project lead}\par}
  \vspace{3.5mm}
  {\renewenvironment{abstract}{\small\noindent\ignorespaces}{\par}%
   \renewcommand{\ourmethod}{\mbox{\textsf{\bfseries 3DCodeVerse}}}%
   \renewcommand{\footnote}[1]{}%
   \begin{abstract}
We introduce \ourmethod{}, a research ecosystem that enables coding agents to create editable and verifiable 3D content at scale.
3DCodeVerse comprises large-scale, high-quality datasets of executable 3D code, a coding harness for spatial reasoning and automated verification, and practical guidance for 3D coding.
\textit{First}, we construct \ourmethod{}\textbf{-2M}, a large-scale curated dataset of standalone 3D executable code.
We build this corpus by collecting open-source data, converting specialized representations, and generating new programs with coding agents.
Each program is packaged with the source files and associated assets required for execution in its declared environment, along with available descriptions and verification records.
\textit{Second}, we develop the \ourmethod{} \textbf{Harness}, a multi-agent directed acyclic graph framework for generating and refining 3D content.
It uses execution checks, spatial measurements, and visual feedback to automatically evaluate and revise generated programs without human feedback during generation.
\textit{Third}, we conduct systematic empirical studies of 3D coding, establishing baselines with fine-tuned open models and deriving practical guidance through component-level analyses of our harness.
Extensive experiments demonstrate that fine-tuning on our dataset significantly improves the 3D coding capabilities of open-source models.
Notably, our 3DCodeVerse harness achieves a 90\% win rate against Claude Design (3D Object) in pairwise human evaluations, when both are utilized by Claude Opus 5.5.%
\end{abstract}
}
  \vspace{3.5mm}
  {\small\raggedright\hypersetup{urlcolor=cardLink}%
  \cardlink{\cvicon}{Project Page}{https://3dcodeverse.com}\\[0.6mm]
  \cardlink{\faGithub}{GitHub}{https://github.com/gaoypeng/3dcodeverse}\\[0.6mm]
  \cardlink{\hficon}{Dataset}{https://huggingface.co/datasets/ilabai/3dcodeverse}\par}
  \end{tcolorbox}
\vspace{2mm}
\begin{center}
\captionsetup{type=figure}
\includegraphics[width=1.0\linewidth]{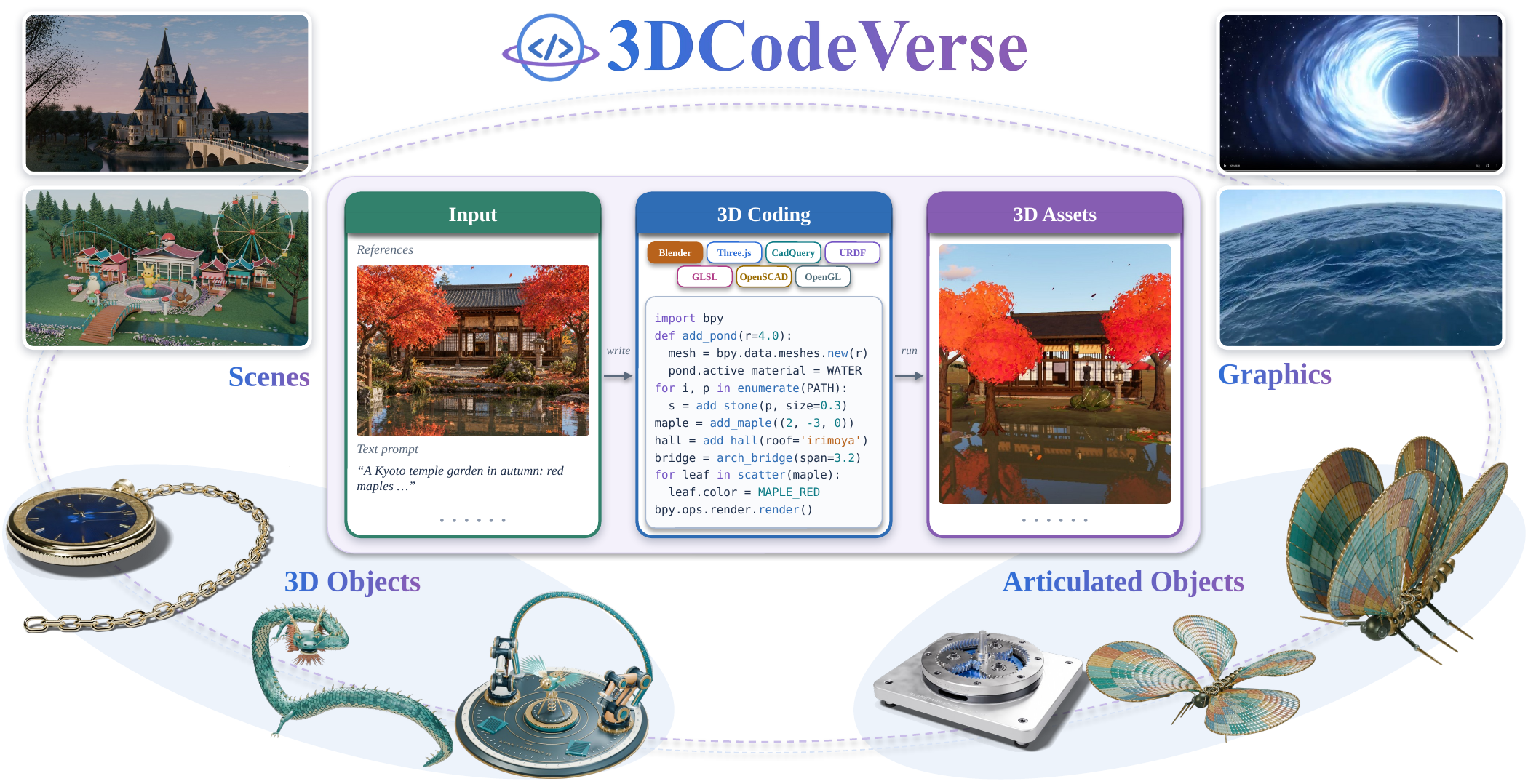}

\vspace{1mm}
\caption{
3DCodeVerse's showcases.
Assets are built \textbf{only by code} (\textit{e.g.}, Blender Python, Three.js, and OpenGL), ranging from 3D objects to articulated objects, scenes, and graphics renderings.
}
\label{fig:teaser}
\end{center}
\vspace{-2mm}

\section{Introduction}
\label{sec:intro}

\nopagebreak
\begin{quote}\centering\small\itshape
``The description of a drawing stored in the Sketchpad system \\ is more than a collection of static drawing parts.''
\par\vspace{2pt}\normalfont\raggedleft
--- \emph{Ivan E. Sutherland, Sketchpad (1963)}
\end{quote}
\nopagebreak

\clearpage

\begin{table}[t]
\centering
\caption{\textbf{Comparison of 3D coding datasets, benchmarks, and harnesses.} Dataset columns show released content, and Graphics marks dedicated shader, material, lighting, or animation data. Sizes follow each source's own unit and are not directly comparable. Harness columns summarize each work's inference-time loop. \pmark{} denotes partial support and \na{} means no new dataset.}
\label{tab:comparison}
\scriptsize
\setlength{\tabcolsep}{2.6pt}
\begin{tabular}{l ccccc rl ccc}
\toprule
 & \multicolumn{5}{c}{\textbf{Dataset}} & \multicolumn{2}{c}{\textbf{Reported size}} & \multicolumn{3}{c}{\textbf{Harness}} \\
\cmidrule(lr){2-6}\cmidrule(lr){7-8}\cmidrule(lr){9-11}
\textbf{Work} & Static & Artic. & Scene & Graphics & \makecell{Code\\Types} & Count & Unit & \makecell{Iterative\\refinement} & \makecell{Visual\\feedback} & \makecell{Geometry\\checks} \\
\midrule
\multicolumn{11}{l}{\textit{Datasets and benchmarks}} \\
Shaders21k \citep{baradad2022procedural}            & \xmark & \xmark & \xmark & \cmark & 1 & 21\,K & \unit{Programs} & \xmark & \xmark & \xmark \\
BlendNet \citep{du2024blenderllm}                   & \cmark & \xmark & \xmark & \xmark & 1 & 8\,K & \unit{Pairs} & \xmark & \xmark & \xmark \\
CAD-Recode \citep{rukhovich2025cadrecode}           & \cmark & \xmark & \xmark & \xmark & 1 & 1\,M & \unit{Programs} & \xmark & \xmark & \xmark \\
MeshCoder \citep{dai2025meshcoder}                  & \cmark & \xmark & \xmark & \xmark & 1 & 1\,M & \unit{Pairs} & \xmark & \xmark & \xmark \\
VLMaterial \citep{li2025vlmaterial}                 & \xmark & \xmark & \xmark & \cmark & 1 & 550\,K & \unit{Materials} & \xmark & \xmark & \xmark \\
BlenderGym \citep{gu2025blendergym}                 & \cmark & \xmark & \cmark & \cmark & 1 & 245 & \unit{Tasks} & \cmark & \cmark & \xmark \\
3DCodeBench \citep{gao20263dcodebench}          & \cmark & \xmark & \xmark & \xmark & 1 & 26\,K & \unit{Scripts} & \cmark & \pmark & \xmark \\
WorldCoder-Bench \citep{lu2026worldcoderbench}      & \xmark & \xmark & \cmark & \cmark & 1 & 2{,}026 & \unit{Tasks} & \xmark & \xmark & \xmark \\
SceneActBench \citep{zhao2026sceneactbench}         & \xmark & \cmark & \cmark & \cmark & 1 & 520 & \unit{Tasks} & \cmark & \cmark & \xmark \\
3DHarnessBench \citep{liu2026_3dharnessbench}    & \cmark & \xmark & \xmark & \xmark & 1 & 100 & \unit{Tasks} & \cmark & \cmark & \pmark \\
\midrule
\multicolumn{11}{l}{\textit{Harnesses and agentic systems}} \\
SceneCraft \citep{hu2024scenecraft}                 & \na & \na & \na & \na & \na & \na & & \cmark & \cmark & \pmark \\
CAD-Assistant \citep{mallis2025cadassistant}        & \na & \na & \na & \na & \na & \na & & \cmark & \cmark & \cmark \\
Articulate-Anything \citep{le2025articulate-anything}& \na & \na & \na & \na & \na & \na & & \cmark & \cmark & \pmark \\
LL3M \citep{lu2025ll3m}                             & \na & \na & \na & \na & \na & \na & & \cmark & \cmark & \xmark \\
LAM \citep{gao2026lam}                              & \xmark & \cmark & \xmark & \xmark & 2 & 2\,K & \unit{Pairs} & \cmark & \cmark & \xmark \\ %
CADCodeVerify \citep{alrashedy2024cadcodeverify}    & \cmark & \xmark & \xmark & \xmark & 1 & 200 & \unit{Prompts} & \cmark & \cmark & \xmark \\
Articraft \citep{zhou2026articraft}                 & \xmark & \cmark & \xmark & \xmark & 2 & 10\,K & \unit{Assets} & \cmark & \xmark & \cmark \\
VIGA \citep{yin2026viga}                            & \xmark & \xmark & \cmark & \cmark & 1 & 27 & \unit{Tasks} & \cmark & \cmark & \xmark \\
\midrule
\textbf{3DCodeVerse (Ours)}                         & \cmark & \cmark & \cmark & \cmark & \textbf{6} & \textbf{2.03\,M} & \unit{Pairs} & \cmark & \cmark & \cmark \\
\bottomrule
\end{tabular}
\end{table}

3D coding offers a promising approach to 3D creation by using executable programs to construct 3D geometry, define articulated motion, compose scenes, and generate graphics effects~\citep{raistrick2023infinigen,hu2024scenecraft,zhou2026articraft}.
Unlike typical 2D generation workflows~\citep{chang2025x-2dgen, ho2020denoising-2dgen, yuan2025magictime-2dgen,yuan2024chronomagic-2dgen}, code-based 3D generation produces structured, executable content that users can edit, reuse, compose, and inspect.
These properties enable scalable, controllable 3D modeling for diverse applications.
Recent advances in language models and coding agents further automate this process, allowing users to create complex 3D content through text instructions~\citep{sun20243dgpt,yin2026viga}.

Despite these advances, shared infrastructure for developing and evaluating 3D coding systems across environments remains limited.
Existing resources address complementary aspects of 3D coding: DeepCAD and BlenderLLM provide supervision for CAD construction and Blender scripting~\citep{wu2021deepcad,du2024blenderllm}; SceneCraft and VIGA refine generated content using visual feedback~\citep{hu2024scenecraft,yin2026viga}; and Articraft combines a specialized SDK with a harness for articulated asset generation~\citep{zhou2026articraft}.
However, these resources remain fragmented across tasks and environments.
As shown in Table~\ref{tab:comparison}, existing datasets primarily focus on static or articulated objects, while harnesses vary in their support for visual assessment and automated corpus construction.

To bridge these gaps, we introduce \ourmethod{} (Figure~\ref{fig:teaser}), an open-source foundational framework for learning and scaling 3D coding through a large-scale executable code corpus and a 3D code generation harness.
We first build \corpus{}, a unified corpus of executable 3D programs spanning static objects, articulated assets, scenes, and graphics.
Rather than simply aggregating existing datasets, we systematically convert heterogeneous 3D representations into reusable source code and remove dependencies on specialized or unavailable libraries.
For example, we translate DeepCAD construction sequences into CadQuery programs and inline Articraft geometry routines while preserving URDF articulation specifications.
Each program is paired with its associated assets, textual descriptions, provenance, and available verification signals.

We further develop \harness{}, a 3D code generation harness for synthesizing and refining complex 3D programs from text, images, or videos.
The harness decomposes multi-component projects into a directed acyclic graph (DAG) of coding and assembly tasks, allowing multiple agents to develop and integrate source files in a structured manner.
Generated programs are executed and rendered, after which an independent judge agent evaluates the outputs against explicit acceptance criteria.
Complementary spatial checks detect structural inconsistencies and prevent visually plausible but geometrically invalid solutions from receiving high scores.
The resulting feedback is then converted into localized repair objectives, enabling repeated execution--evaluation--revision cycles without human intervention.
Through this iterative process, the harness supports \emph{inference-time scaling} to complex projects containing thousands of lines of code.

Importantly, \harness{} serves not only as an inference system but also as a data-generation engine.
By recording successful programs, failed attempts, targeted repairs, and preference comparisons, its execution traces provide supervision for both supervised fine-tuning and preference learning.
Together, \corpus{} and \harness{} establish not only a reusable infrastructure for 3D coding, but also a closed loop connecting data construction, program synthesis, verification, refinement, and model improvement.
We further support this ecosystem with open models and reproducible training recipes.
Fine-tuning on \corpus{} substantially improves executable 3D code generation, reaching execution rates of 84.9\% with a 9B model and 92.9\% with a 27B model on \bench{}.
Moreover, \harness{} consistently improves generation quality across coding agents, yielding gains of 15.5--29.1 points over one-shot generation.

\section{Related Work}
\label{sec:related}

\paragraph{Datasets and Benchmarks for 3D Coding.}
Learning to generate 3D content as code benefits from supervision over its construction procedures.
DeepCAD and Fusion~360 Gallery expose CAD construction sequences, and Text2CAD connects such sequences to natural language~\citep{wu2021deepcad,willis2021fusion360,khan2024text2cad}.
This supervision has expanded to executable CadQuery and Blender programs for reconstruction and language-conditioned generation through BlendNet, CAD-Recode, CAD-Coder, and MeshCoder~\citep{du2024blenderllm,rukhovich2025cadrecode,guan2025cadcoder,dai2025meshcoder}.
Complementary procedural resources broaden the content represented by code: Infinigen and its extensions generate assets and scenes~\citep{raistrick2023infinigen,raistrick2024infinigenindoors,joshi2025infinigensim}, while Shaders21k provides programs for procedural image synthesis~\citep{baradad2022procedural}.
Existing benchmarks assess the resulting code generation, editing, and physics capabilities~\citep{alrashedy2024cadcodeverify,gu2025blendergym,gao20263dcodebench, shen20264dcodebench}.
However, our paper focuses on scaling executable supervision across these heterogeneous coding environments.
Our \ourmethod{}-2M provides a million-scale training corpus spanning CAD, Blender, shaders, and browser graphics, with programs that are self-contained within their respective target runtimes.
Its construction combines curated existing sources with automated program generation and execution-based filtering, enabling continued data expansion for training open 3D coding models.

\paragraph{Harness Engineering for 3D Coding.}
Execution and feedback allow 3D coding agents to inspect and revise the content produced by their programs.
Systems such as 3D-GPT, SceneCraft, and LL3M establish language-driven planning and program generation for 3D modeling~\citep{sun20243dgpt,hu2024scenecraft,lu2025ll3m}.
Executing and rendering these programs enables visual refinement, as explored by BlenderAlchemy, CADCodeVerify, and VIGA~\citep{huang2024blenderalchemy,alrashedy2024cadcodeverify,yin2026viga}.
Beyond rendered appearance, Articraft provides structured asset validation through a specialized SDK, while Bioinspired123D connects script quality control with a graph-based agentic workflow~\citep{luu2026bioinspired123d,zhou2026articraft}.
3DHarnessBench further studies how the observation and tool interfaces available to an agent affect 3D reconstruction~\citep{liu2026_3dharnessbench}.
In contrast, our \ourmethod{} harness provides a reusable execution and verification framework across coding backends, jointly supporting inference-time refinement and scalable training data curation.
Its multi-agent directed acyclic graph coordinates execution, rendering, spatial measurements, and structural checks across static and articulated objects, scenes, and visual effects.
The same framework equips coding agents with visual and programmatic feedback and serves as an automated data engine that generates, refines, and filters training programs without human feedback in the curation loop.

\section{\corpus{}: A large-scale 3D coding dataset}
\label{sec:corpus}

\noindent\textbf{Task Definition of 3D Coding.}
We define \emph{3D coding} as writing a program $f$ that constructs 3D content satisfying a condition $c$, comprising a text request and optional reference images.
The program is standalone, bundling the source files and assets needed to run in a declared environment $e$ such as Blender (\texttt{bpy}) or CadQuery.
Analogous to compiling source code, executing $f$ in $e$ yields
\begin{equation}
  (a,\ell)=\mathcal{E}_e(f), \qquad x=\mathbb{I}[a\neq\bot],
  \label{eq:execute}
\end{equation}
where $a$ is the artifact (a mesh, an articulated asset, a scene, or an animation), $\ell$ is the execution feedback, and $\bot$ denotes failure.
Unlike conventional code generation, success requires not only executability ($x=1$) but also high-quality 3D content.
Quality is measured by a task-specific metric $\mathcal{D}(a,a^\star)$ when a reference $a^\star$ exists, and otherwise against $c$ using programmatic measurements and renders $o=\mathcal{R}(a)$.

\noindent\textbf{3D Coding with LLMs and Coding Agents.}
A generation policy $\pi$ produces the program.
A single-shot LLM writes $f^{(0)}=\pi(c;e)$ directly.
A coding agent instead iterates like a developer in a compile--debug loop, revising its code based on execution feedback and renders:
\begin{equation}
  f^{(t)}=\pi\big(c,\{f^{(k)},\ell^{(k)},o^{(k)}\}_{k<t};e\big),
  \label{eq:agent}
\end{equation}
where $(a^{(t)},\ell^{(t)})=\mathcal{E}_e(f^{(t)})$ and $o^{(t)}=\mathcal{R}(a^{(t)})$.
After $T$ refinement rounds, the agent returns $f^{(T)}$, so the single-shot setting is the case $T=0$.
The harness implements $\mathcal{E}_e$ and $\mathcal{R}$: agents call them as tools, and all final programs are evaluated with them under the same protocol.

\begin{figure}[!t]
\centering
\includegraphics[width=\linewidth]{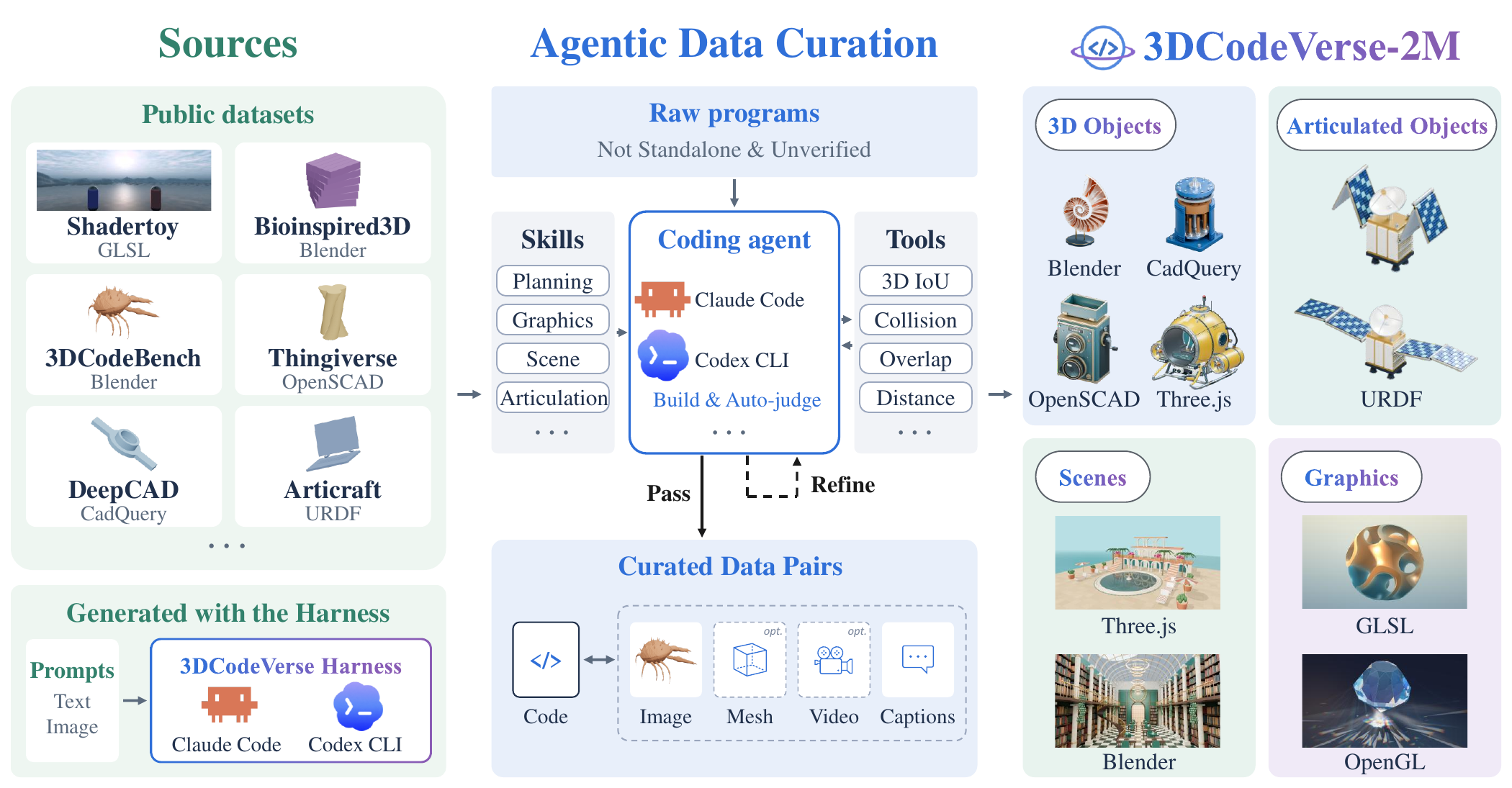}
\caption{
\textbf{Curation and composition of \corpus{}.}
Left: We collect programs from public datasets and generate more with our harness.
Middle: Our 3D coding harness, equipped with modeling skills and inspection tools (e.g., 3D IoU, collision checks), builds each program, judges the output, and refines it until it passes.
Each verified program is paired with rendered images and captions, plus meshes and videos when available.
Right: The resulting corpus covers 3D objects, articulated objects, scenes, and graphics across multiple programming languages and engines.
}
\label{fig:dataset-overview}
\end{figure}
\noindent\textbf{Data Sources and Composition.}
{
Figure~\ref{fig:dataset-overview} shows representative sources; Appendix~\ref{app:corpus-sources} summarizes the full collection.
Public resources include Bioinspired3D~\citep{bioinspired3d}, \bench{}~\citep{gao20263dcodebench}, Thingiverse, Shadertoy~\citep{shadertoy}, CodePen animations (Animation2Code), and multi-file Three.js/React Three Fiber projects.
We convert DeepCAD sequences~\citep{wu2021deepcad} and Articraft geometry~\citep{zhou2026articraft} to CadQuery, retaining Articraft's URDF.
The ProcTHOR~\citep{deitke2022procthor} subset pairs scene specifications with Blender/Three.js reconstructions.
Blender and Three.js coding-agent subsets contribute generated objects and scenes; We further synthesize and archive new programs by running \harness{} on text and image prompts (\S\ref{sec:harness}).
}

\noindent\textbf{Agentic Data Curation and Toolkits.}
Coding agents curate programs with reusable \emph{skills} (Appendix~\ref{app:harness-skills}) that package authoring conventions, code examples, and repair guidance.
Given a raw program $f_{\mathrm{raw}}$ that relies on specialized APIs, an agent produces a standalone program $f=\mathcal{N}_e(f_{\mathrm{raw}})$ for environment $e$ by translating those APIs, inlining required routines, and bundling sources and assets; DeepCAD is instead converted by a deterministic script (Appendix~\ref{app:corpus-sources}).

Executing $f$ yields $(a,\ell)=\mathcal{E}_e(f)$ (Eq.~\ref{eq:execute}), and inspection \emph{tools} return measurements $\mathcal{M}_e(a)$: part inventories and bounding boxes flag missing or misplaced components, contact and collision checks flag gaps and interpenetration, including across sampled joint poses, and placement and frame probes check physical support, invalid renders, and temporal defects.
Findings are reported per part in the authoring frame (Appendix~\ref{app:harness-tools}), and execution errors or failed checks return the program to the agent for targeted repair.
Conversion fidelity is measured by voxel IoU for DeepCAD and Chamfer distance for Articraft.
Coverage thus varies by source: collected and converted programs are checked by execution and rendering where a runtime is available, whereas harness-generated programs also carry gate and judge results as quality tiers.

A VLM captions programs from code and renders $o=\mathcal{R}_e(a)$ with three conditions $C=\{c^{\mathrm{inst}},c^{\mathrm{det}},c^{\mathrm{proc}}\}$: an instruction, a visual description, and a construction procedure, each usable as $c$ in Eq.~\ref{eq:execute}.
Each record $s=(f,e,C,o,m)$ stores provenance, license, and available verification results in $m$, keeping execution, geometric, and visual checks separate; source files, pre-conversion programs, and URDF files are retained, and meshes and videos are linked when available.

\section{\harness: A Data Engine for 3D Coding}
\label{sec:harness}

\begin{figure}[t]
\centering
\includegraphics[width=0.95\linewidth]{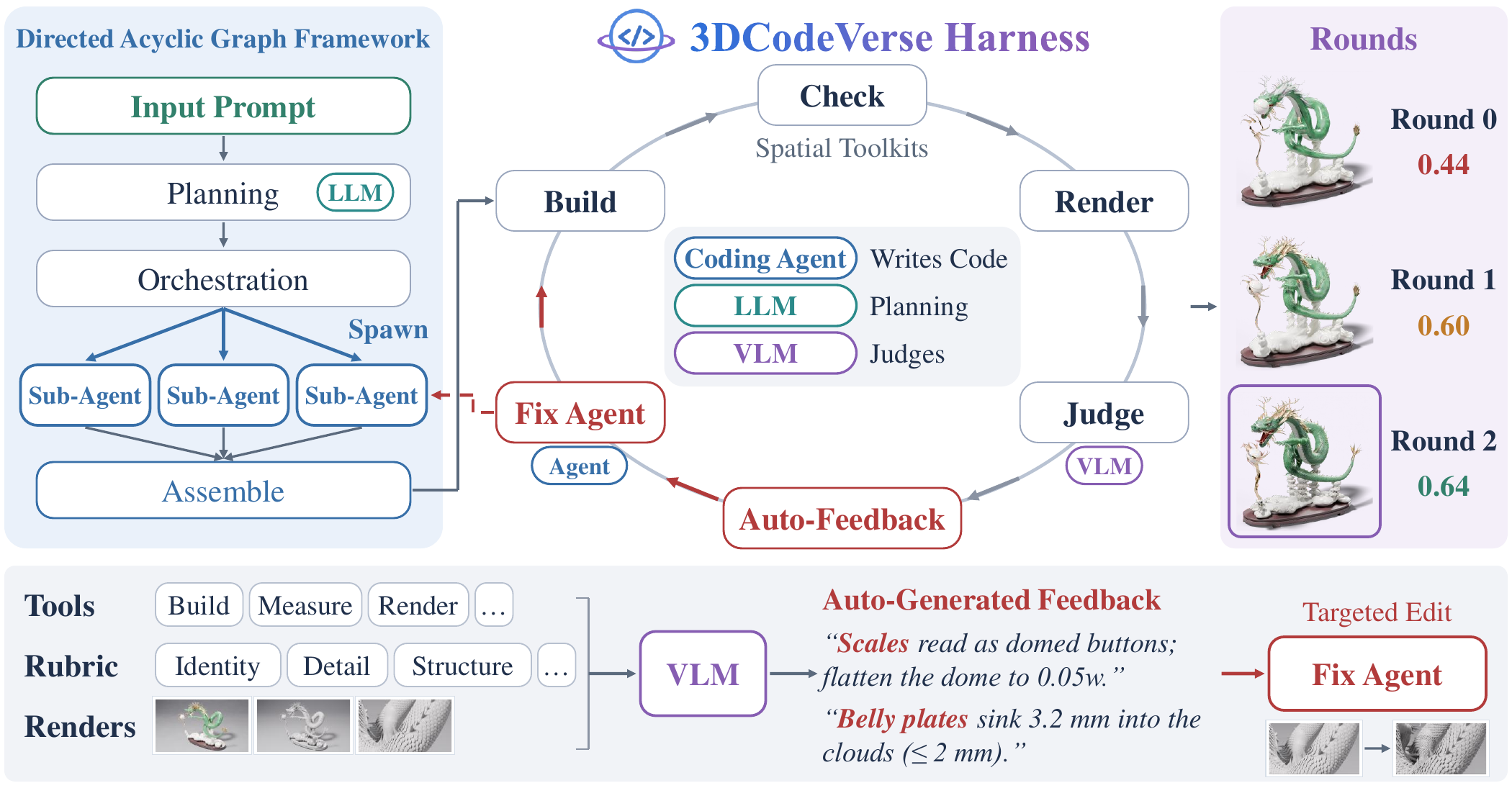}
\caption{
\textbf{Overview of the \harness{}.}
Given an input prompt, an LLM planner defines the target asset, and a DAG coordinates sub-agents to code and assemble a 3D program (left).
Execution checks, spatial measurements, and rubric-guided VLM judgments provide automated feedback (center) for targeted code revisions by a fix agent (bottom).
The right panel illustrates successive refinement rounds, with the highest-scoring program selected as the final output.
}

\label{fig:pipeline}
\end{figure}

Generating high-quality 3D programs often requires iterative refinement, but reliance on human feedback makes this process difficult to scale.
To address this bottleneck, we develop \harness{}, a multi-agent framework that automates 3D code generation and refinement for data curation and agent-driven content creation (Figure~\ref{fig:pipeline}).
It coordinates coding and assembly through a DAG and combines execution checks, spatial tools, and rubric-guided VLM judgments to identify defects and guide targeted revisions without human feedback.
Because the reliability of this feedback depends on how we configure its prompts, tools, and rubrics, we further refine the judge and generation workflow through an alternating development-time adversarial loop.

\noindent\textbf{Multi-Agent Directed Acyclic Graph (DAG) Framework.}
Given a condition $c$ and a target execution environment $e$, an LLM planner first produces $P=\mathcal{A}_{\mathrm{plan}}(c;e)$, specifying the target components, dimensions, materials, spatial relations, and acceptance criteria.
The harness translates $P$ into a DAG $\mathcal{G}=(V, E)$ of coding and assembly tasks, with edges encoding prerequisites.
Within this graph, explicit file ownership allows independent coding tasks to be carried out concurrently by sub-agents.
Once their prerequisites are satisfied, assembly tasks integrate the resulting code, ultimately producing the initial program $f^{(0)}$ for subsequent evaluation and feedback-driven refinement.

\paragraph{Automated Feedback and Iterative Refinement.}
At each round $t$, the harness executes the current program $f^{(t)}$ and collects three types of feedback:
\begin{enumerate}[label=\textcircled{\small\arabic*},leftmargin=*,itemsep=1pt,topsep=2pt]
  \item \textbf{Execution feedback.}
  Build and runtime logs $\ell^{(t)}$.
  If execution fails, no artifact is produced, and this is the only feedback.
  \item \textbf{Measured feedback.}
  Measurement tools $\mathcal{M}_e$ check the artifact $a^{(t)}$ for dimensions, contacts, collisions, kinematics, and placement, and return component-level violations with numerical correction hints.
  \item \textbf{Visual feedback.}
  A VLM judge inspects multi-view renders $\mathcal{R}_e(a^{(t)})$ against the input $c$ and the acceptance criteria in $P$, producing defect evidence, criterion scores, and proposed revisions: $j^{(t)}=\mathcal{A}_{\text{judge}}(c,P,o^{(t)})$, where $o^{(t)}=(\ell^{(t)},\mathcal{M}_e(a^{(t)}),\mathcal{R}_e(a^{(t)}))$.
\end{enumerate}
The score $s^{(t)}$ combines the weighted criteria and defect penalties in $j^{(t)}$, caps it with measured violations and unmet requirements, and vetoes measurements that rule out visual defect claims.
All feedback is converted into revision tasks $\Delta^{(t)}$, prioritized in the order listed above, and a fix agent applies them: $f^{(t+1)}=\mathcal{A}_{\text{code}}(c,P,f^{(t)},\Delta^{(t)};e)$.

\noindent\textbf{Adversarial Loop Engineering for Harness Optimization.}
The preceding loop uses automated feedback in place of per-round human inspection and revision guidance.
Because this feedback's reliability depends on how prompts, tools, and rubrics are configured, we alternate between refining the judge and the generation workflow during development.
Let $\theta_J$ denote the judge's evidence presentation, rubric, response schema, and aggregation rules, and $\theta_H$ the harness's prompts, contracts, skills, tools, and refinement policy.
\emph{Judge refinement} first fixes the generation workflow and repeatedly evaluates a bank $\mathcal{B}$ of stored artifacts.
Using the same artifacts separates changes in judge behavior from variability in generation.
We revise view selection, evidence presentation, and defect checklists, and score aggregation to reduce judgment variability and contradictions with measured evidence.
The judge is required to report observations before scores and does not receive the generator's reasoning.
We summarize the consistency objective as:
\begin{equation}
  \mathcal{L}_J(\theta_J)
  =
  \mathbb{E}_{(a,o)\sim\mathcal{B}}
  \Bigl[
    \operatorname{Var}_r\,s^{(r)}(o;\theta_J)
    +
    \lambda\,\rho\bigl(
      j(o;\theta_J),\mathcal{M}_e(a)
    \bigr)
  \Bigr],
  \label{eq:judge-opt}
\end{equation}
where $r$ indexes repeated judgments and $\rho$ counts claims contradicted by applicable measurements.
Preserving sensitivity to differences in artifact quality remains an additional requirement beyond this consistency objective.
With the revised judge fixed, \emph{harness refinement} uses its feedback to guide changes to the generation workflow.
Candidate designs are compared on matched prompts, balancing judged output quality against observable workflow failures:
\begin{equation}
  \mathcal{U}_H(\theta_H;\theta_J)
  =
  \mathbb{E}_{c}
  \Bigl[
    s\bigl(\hat{f}(c;\theta_H);\theta_J\bigr)
  \Bigr]
  -
  \mu\,\mathcal{L}_{\mathrm{fail}}(\theta_H),
  \label{eq:harness-opt}
\end{equation}
where $s(\hat{f};\theta_J)$ denotes the score obtained by executing and judging the selected program $\hat{f}$, and $\mathcal{L}_{\mathrm{fail}}$ summarizes failures such as invalid plans, retried tool calls, and failed builds.
The weights $\lambda,\mu\geq0$ express the respective trade-offs.

\FloatBarrier

\section{Experiments}
\label{sec:experiments}

\subsection{Results of 3DCodeVerse Harness}
\label{sec:exp-harness}

\noindent\textbf{Evaluation setup.}
We compare \ourmethod{} against models with their native harnesses and Claude Design on 100 text-to-3D object prompts.
We further compare against native harnesses on articulated-object generation, 50 scene prompts, and 50 graphics prompts.
We evaluate generation quality using three metrics.
\emph{Human Preference} is the fraction of anonymized side-by-side comparisons won according to human evaluators, with ties split equally.
\emph{VLM Preference} is the fraction of the same comparisons won according to a VLM judge (Gemini~3.1~Pro), with ties split equally.
\emph{Judge Score} is Gemini~3.1~Pro's overall rubric score, combining weighted criteria with defect penalties and score caps (0--100; failed generations receive a score of zero).
Following 3DCodeBench~\citep{gao20263dcodebench}, we additionally report SigLIP-2~\citep{tschannen2025siglip2} text--image cosine similarity averaged over four views.
The evaluation judge is described in Appendix~\ref{app:eval-judge}, and further evaluation details are in Appendix~\ref{app:eval-protocol}.

\begin{figure}[t]
\centering
\includegraphics[width=0.95\linewidth]{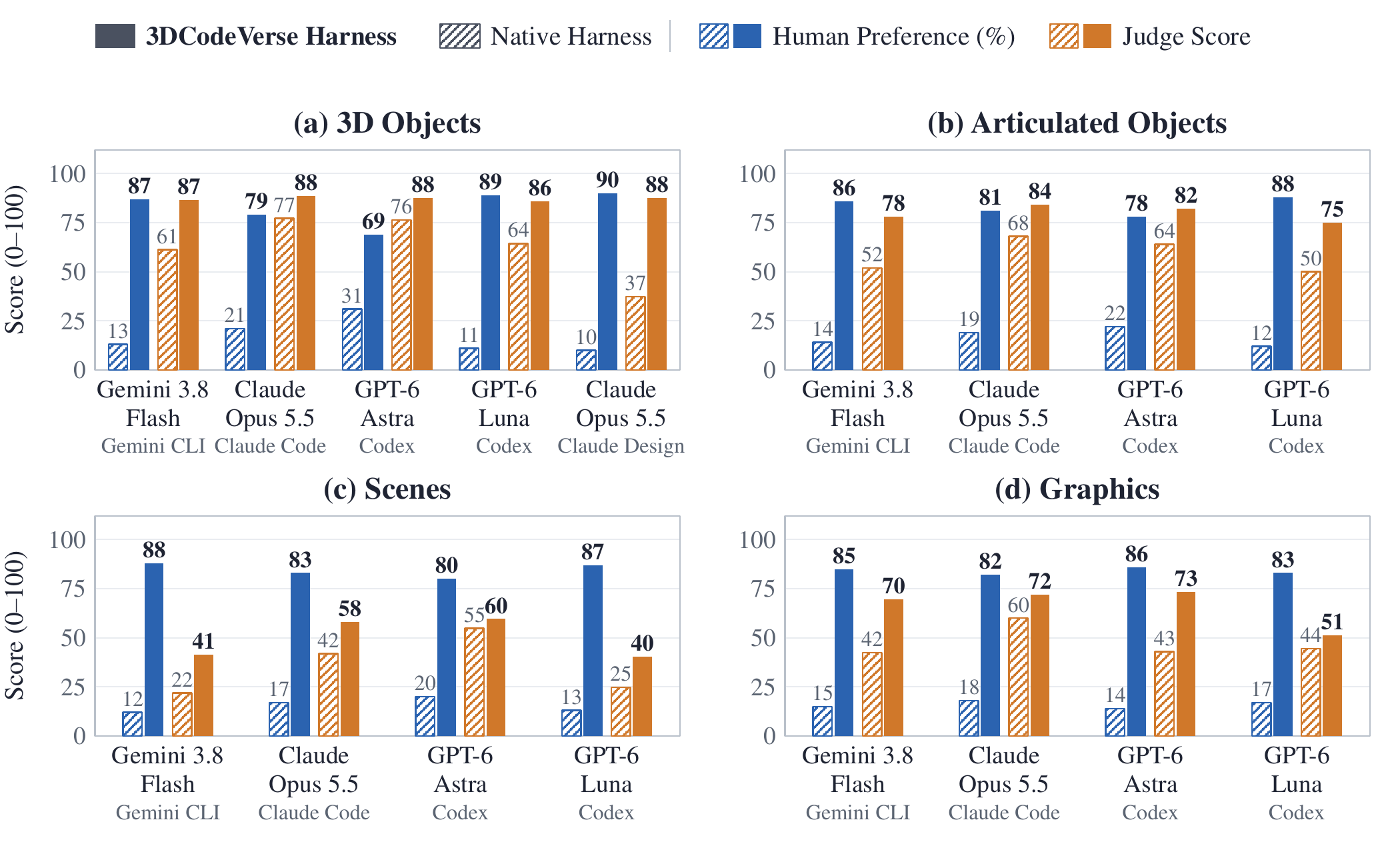}
\caption{
\textbf{Main results of \harness{} across four 3D coding tasks.}
Panels show (a) 3D objects, (b) articulated objects, (c) scenes, and (d) graphics.
Blue bars report Human Preference (\%), and orange bars report Judge Score (0--100; failed generations receive a score of zero).
Hatched bars denote each backbone's native harness, while solid bars denote \harness{}.
The final group in (a) compares Claude Design with \harness{}, both using Claude Opus~5.5.
}
\label{fig:main-results}
\end{figure}

\noindent\textbf{Main Results.}
Figure~\ref{fig:main-results} shows a consistent pattern across backbones and tasks: outputs from the native harness are rarely preferred, while the same backbone within \harness{} wins most comparisons.
On 3D objects, \emph{Human Preference} for \harness{} ranges from 69\% to 89\%, while \emph{Judge Score} improves by 11--25 points.
The largest gains occur for the weaker models, including Gemini~3.8~Flash (61 to 87) and GPT-6~Luna (64 to 86).
Against Claude Design, which uses Claude Opus~5.5, \harness{} with the same backbone is preferred in 90\% of human comparisons and achieves a Judge Score of 88 versus 37.
The same pattern extends to articulated objects, scenes, and graphics, where Human Preference reaches 78--88\%.
The agreement between human preference and rubric-based scores suggests that the gains are not specific to the VLM rubric used during refinement.

\subsection{Fine-tuning Open Models on \corpus{}}
\label{sec:exp-sft}

\providecommand{\tbd}{\textcolor{black!35}{\ldots}}
\begin{table}[!b]
\centering
\caption{\textbf{Fine-tuning with \corpus{} makes open models executable.}
Execution rate (\%) before and after LoRA fine-tuning (training-set size in parentheses); $\Delta$ is the change and the better value of each pair is in bold.
A program counts when it runs in the dialect's runtime and produces a result (GLSL: the shader compiles).
All models are decoded at $T{=}0.7$ with one seed on the same fixed test sets.}
\label{tab:sft}
\small
\setlength{\tabcolsep}{5pt}
\begin{tabular}{@{}l cccccc@{}}
\toprule
\textbf{Model} & \bench{} & Blender & CadQuery & OpenSCAD & GLSL & Three.js \\
& (212) & (103) & (200) & (50) & (200) & (40) \\
\midrule
Qwen3.5-9B & 0.0 & 1.9 & 10.5 & 50.0 & 33.0 & 72.5 \\
\quad + SFT  & \textbf{84.9} & \textbf{90.3} & \textbf{98.5} & \textbf{86.0} & \textbf{56.0} & \textbf{74.5} \\
\quad {\footnotesize$\Delta$} & {\footnotesize\textcolor{green!40!black}{+84.9}} & {\footnotesize\textcolor{green!40!black}{+88.4}} & {\footnotesize\textcolor{green!40!black}{+88.0}} & {\footnotesize\textcolor{green!40!black}{+36.0}} & {\footnotesize\textcolor{green!40!black}{+23.0}} & {\footnotesize\textcolor{green!40!black}{+2.0}} \\
\midrule
Qwen3.8-27B & 51.9 & 74.8 & 65.5 & 80.0 & 57.0 & 72.5 \\
\quad + SFT + DPO & \textbf{92.9} & \textbf{93.2} & \textbf{98.5} & \textbf{90.0} & \textbf{71.5} & \textbf{92.5} \\
\quad {\footnotesize$\Delta$} & {\footnotesize\textcolor{green!40!black}{+41.0}} & {\footnotesize\textcolor{green!40!black}{+18.4}} & {\footnotesize\textcolor{green!40!black}{+33.0}} & {\footnotesize\textcolor{green!40!black}{+10.0}} & {\footnotesize\textcolor{green!40!black}{+14.5}} & {\footnotesize\textcolor{green!40!black}{+20.0}} \\
\bottomrule
\end{tabular}
\end{table}

\begin{figure}[!t]
\centering
\includegraphics[width=0.9\linewidth]{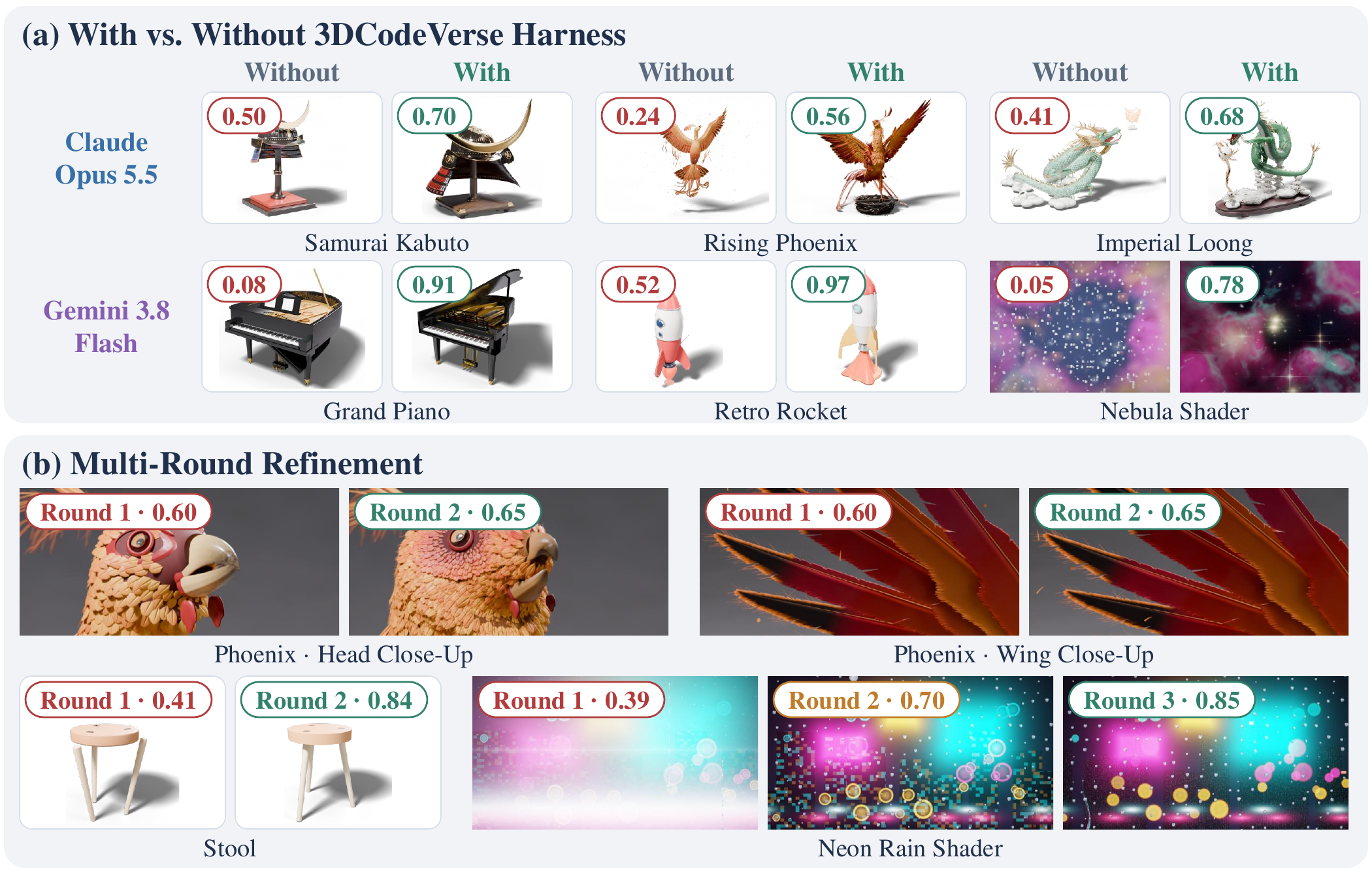}
\caption{
\textbf{Qualitative comparisons and automated refinement.}
\textbf{(a)}~Outputs for identical prompts and backbones using the native harness (\emph{Without}) versus \harness{} (\emph{With}).
\textbf{(b)}~Selected rounds of automated refinement for objects and graphics.
Numbers report rubric scores on a $[0,1]$ scale and are comparable within each pair or refinement sequence.
}
\label{fig:qualitative}

\vspace{0.8em}
{\captionsetup{type=table}\centering
\caption{\textbf{Fidelity to the reference before and after fine-tuning.}
Same models, decoding, and layout as Table~\ref{tab:sft}.
\emph{F}: F@0.05 between generated and reference meshes.
\emph{DINO}: DINOv2-L similarity between four-view renders of generated and reference meshes.
}
\label{tab:sft-fidelity}
\small
\setlength{\tabcolsep}{5pt}
\begin{tabular}{@{}l cc ccc@{}}
\toprule
& \multicolumn{2}{c}{\bench{}} & \multicolumn{3}{c}{F@0.05 per dialect} \\
\cmidrule(lr){2-3}\cmidrule(lr){4-6}
\textbf{Model} & F & DINO & Blender & CadQuery & OpenSCAD \\
\midrule
Qwen3.5-9B & .000 & .000 & .008 & .030 & .261 \\
\quad + SFT & \textbf{.317} & \textbf{.266} & \textbf{.693} & \textbf{.308} & \textbf{.329} \\
\quad {\footnotesize$\Delta$} & {\footnotesize\textcolor{green!40!black}{+.317}} & {\footnotesize\textcolor{green!40!black}{+.266}} & {\footnotesize\textcolor{green!40!black}{+.685}} & {\footnotesize\textcolor{green!40!black}{+.279}} & {\footnotesize\textcolor{green!40!black}{+.069}} \\
\midrule
Qwen3.8-27B & .214 & .211 & .439 & .159 & \textbf{.415} \\
\quad + SFT + DPO & \textbf{.374} & \textbf{.298} & \textbf{.798} & \textbf{.234} & .416 \\
\quad {\footnotesize$\Delta$} & {\footnotesize\textcolor{green!40!black}{+.160}} & {\footnotesize\textcolor{green!40!black}{+.088}} & {\footnotesize\textcolor{green!40!black}{+.359}} & {\footnotesize\textcolor{green!40!black}{+.075}} & {\footnotesize\textcolor{green!40!black}{+.001}} \\
\bottomrule
\end{tabular}
}
\end{figure}

\paragraph{Setup.}
We fine-tune Qwen3.5-9B and Qwen3.8-27B with LoRA~\citep{hu2022lora} in LLaMA-Factory~\citep{zheng2024llamafactory} on mixtures drawn from \corpus{} (Appendix~\ref{app:training}) and compare the resulting models with their respective pretrained checkpoints.
Evaluation spans 805 fixed tasks across six suites: 212 text-to-Blender tasks from \bench{}~\citep{gao20263dcodebench} and five test splits drawn from \corpus{}: Blender (103 procedural objects), CadQuery (200 mechanical parts from DeepCAD and Articraft), OpenSCAD (50 Thingiverse models), GLSL (200 Shadertoy shaders), and Three.js (40 single-object scenes).
Each task pairs a natural-language request with a reference program.
Executing the reference program produces the evaluation target: a mesh for the four 3D suites, a rendered frame for GLSL, and a page screenshot for Three.js.
All models receive the same request and per-dialect system prompt without examples, are decoded at $T{=}0.7$ with one seed, and have every output executed in the dialect's pinned runtime (Appendix~\ref{app:eval-protocol}).
We report the execution rate (Table~\ref{tab:sft}) and fidelity to the target (Table~\ref{tab:sft-fidelity}): F@0.05 between the generated and reference meshes, and DINOv2-L similarity between clay renders of the two meshes from four fixed views; failed programs score zero on both.

\paragraph{Fine-tuning substantially improves executability.}
Before fine-tuning, none of the programs generated by Qwen3.5-9B on \bench{} execute successfully; most fail immediately by calling scene-clearing operators unavailable in Blender~5.
After fine-tuning on \corpus{}, execution rises to \textbf{84.9\%} on \bench{}, 90.3\% on Blender, and 98.5\% on CadQuery, with additional gains of 36.0 and 23.0 points on OpenSCAD and GLSL (Table~\ref{tab:sft}).
Qwen3.8-27B starts from a stronger baseline but shows the same trend: fine-tuning raises execution on \bench{} from 51.9\% to \textbf{92.9\%} (+41.0 points), on Blender to 93.2\% (+18.4), and on CadQuery to 98.5\% (+33.0).

\subsection{Qualitative Analysis}
\label{sec:exp-qualitative}
For matched prompts and backbones, Figure~\ref{fig:qualitative}(a) shows that \harness{} produces more complete structures and finer details, such as the phoenix's legs and nest and the piano's interior.
Figure~\ref{fig:qualitative}(b) shows how successive refinement rounds correct specific defects, improving the phoenix's facial feathers, attaching the stool legs to the seat, and reducing block artifacts in the neon shader.
Together, these examples show that automated feedback can correct both structural and appearance defects without additional human instructions.

\FloatBarrier

\clearpage
\section{Applications of 3D coding}
\label{sec:applications}

\begin{figure}[!t]
\centering
\includegraphics[
    width=\linewidth,
    height=0.30\textheight,
    keepaspectratio
]{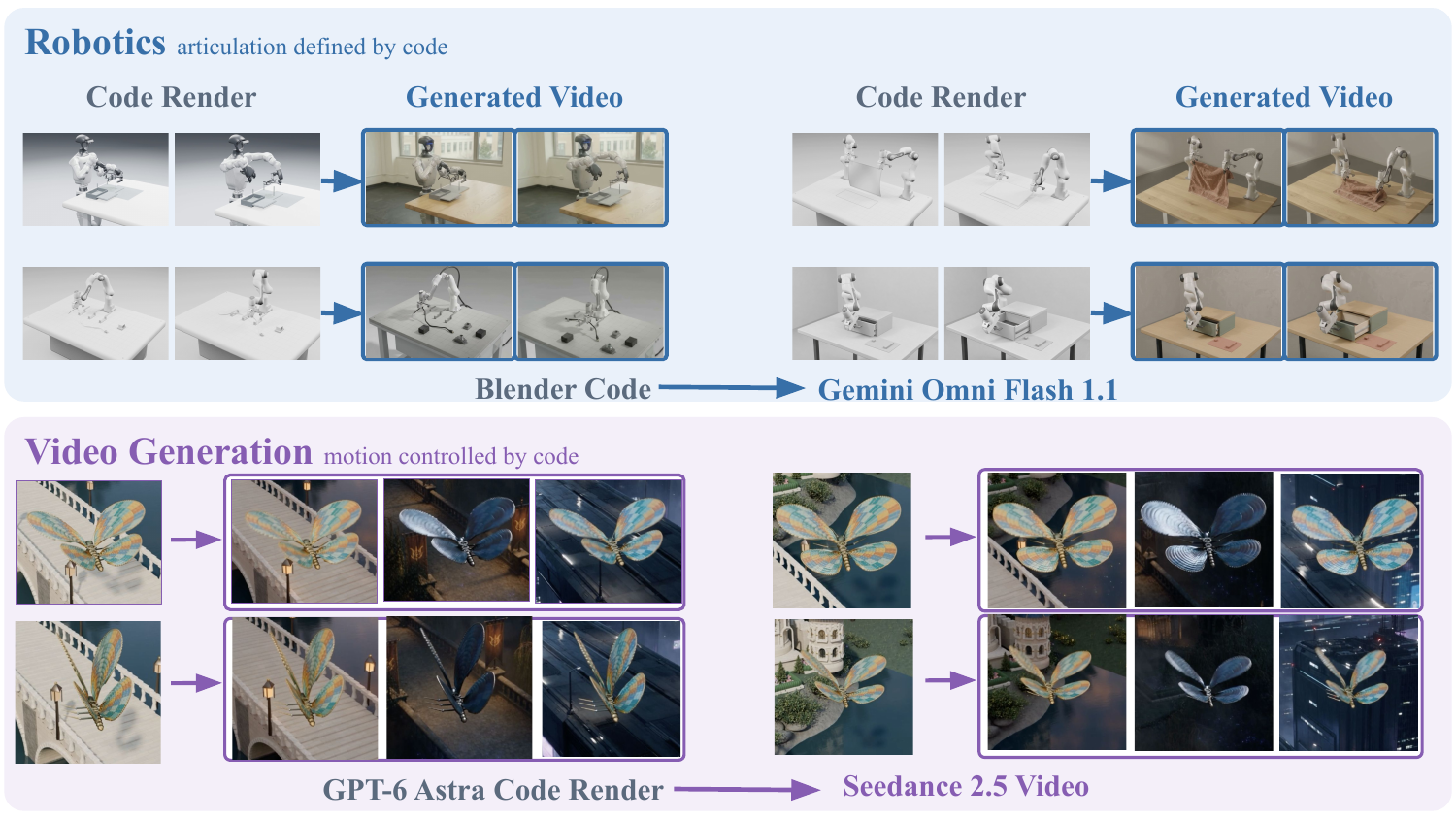}
\caption{
\textbf{Applications.}
Code renders from our harness guide video generation with Gemini Omni Flash~1.1 for robot manipulation scenes (top) and Seedance~2.5 for a GPT-6 Astra scene (bottom).
}
\label{fig:applications}
\end{figure}

\textbf{Robotics.}
3D coding enables the construction of articulated assets and scenes with explicit geometry and physical constraints for simulated robot tasks.
Programs encode rigid-link geometry, joint axes, and joint limits as editable parameters, supporting programmatic checks and controlled variation of object dimensions, articulation, and initial poses.
These task instances can be executed in simulations to generate robot interactions and observations.
To provide a preliminary evaluation, we import an articulated asset validated and exported with 3DCodeVerse tools into MuJoCo and compare rendered observations with frozen UWM~\citep{zhu2025uwm} predictions across 12 cases spanning two assets, two contact states, and three control sequences, using complete scene state for rendering and image history with matched robot commands for UWM.
Code renders and their UWM VAE reconstructions each pass a fixed door-and-handle geometry check in 12/12 outputs, versus 7/36 UWM predictions.
These results show that 3D code can help generate articulated assets and scenes for robotics simulation with explicit geometry and physical constraints.

\textbf{Video generation.}
3D coding provides an editable interface for controlling the spatial and temporal conditions of video generation.
Programs specify scene layout, object articulation and motion, and camera trajectories.
We render generated 3D scenes using programs and harness them as reference videos, using the full sequences to condition Gemini Omni Flash~\citep{google2026geminiomni} and Seedance 2.5~\citep{bytedance2026seedance25}.
The rendered videos supply spatial and temporal guidance, while the video models are used to refine visual appearance.
We present paired reference and generated frames to examine how the specified layout, object motion, and camera trajectories are preserved.
This application connects explicit scene programs with video synthesis.

\section{Conclusion}
\label{sec:conclusion}

In this paper, we present \ourmethod{}, an open ecosystem that provides a shared foundation for 3D coding through datasets, a reusable harness, and open model baselines.
The dataset brings together programs across coding environments with descriptions, associated assets, and available quality records.
The harness combines execution, spatial checks, and visual judgment to support iterative generation without human feedback during a run, while retaining the resulting programs and revisions for data curation.
Experiments demonstrate the utility of the data for training open models and the benefits of the harness for existing coding agents.
We will release the datasets, harness, models, and training recipes to support further research on 3D generation through code.

\clearpage

\bibliography{references}
\bibliographystyle{iclr2027_conference}

\appendix

\renewcommand{\topfraction}{0.92}
\renewcommand{\bottomfraction}{0.6}
\renewcommand{\textfraction}{0.06}
\renewcommand{\floatpagefraction}{0.6}
\setcounter{topnumber}{3}
\setcounter{totalnumber}{4}
\raggedbottom

\clearpage
\section*{Appendix}
\label{app:contents}

\startcontents[appendix]
\begingroup
\footnotesize
\printcontents[appendix]{l}{1}{\setcounter{tocdepth}{2}}
\endgroup

\clearpage
\section{The \harness{}: Design and Implementation}
\label{app:harness}

\subsection{Components and model roles}
\label{app:harness-overview}

\begin{figure}[!tbp]
\centering
\includegraphics[width=\linewidth]{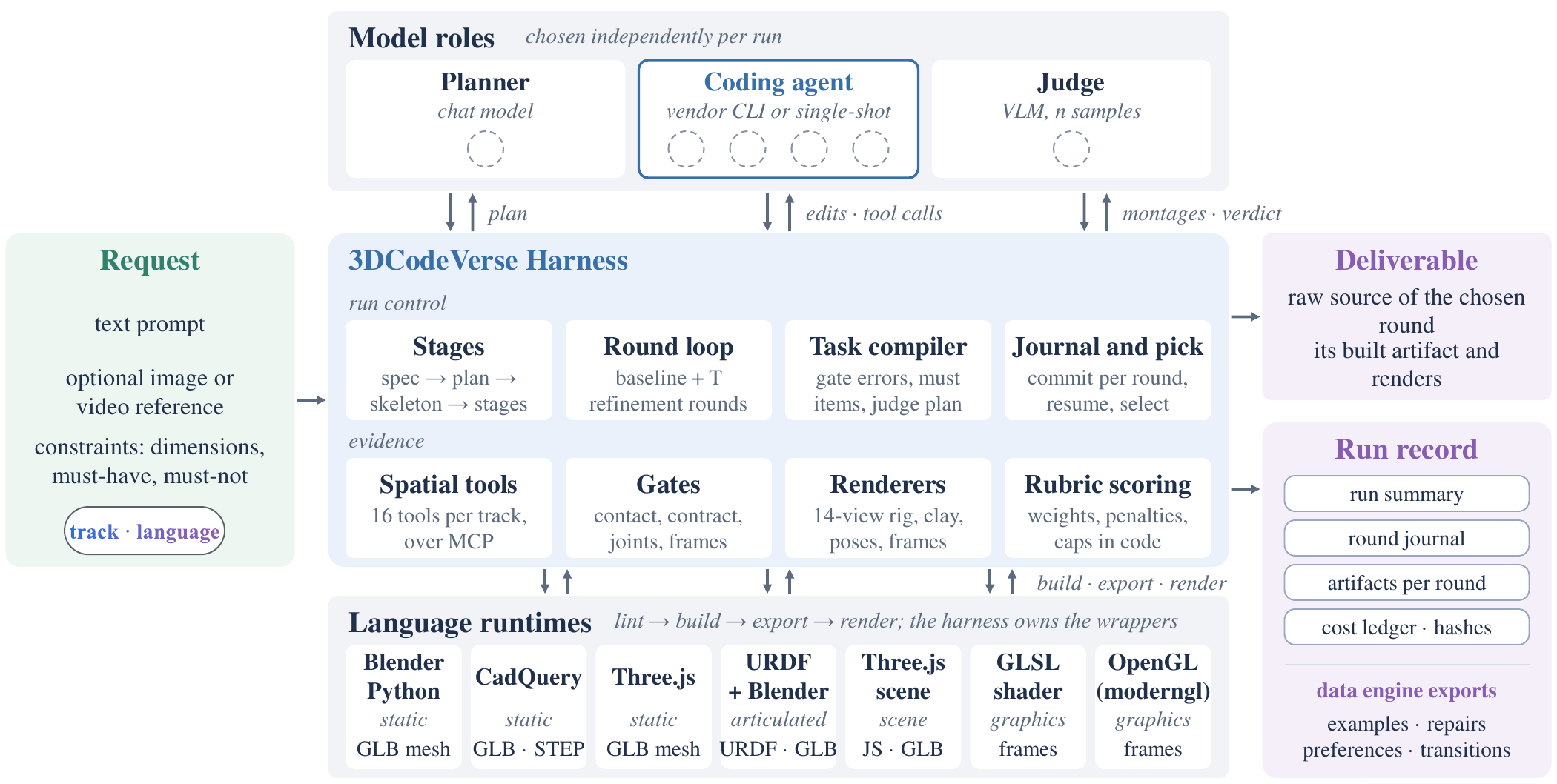}
\caption{\textbf{Components and model roles.}
A planner, a judge, and a coding agent are configured modularly per run; the harness orchestrates stage progression, iterative refinement, and candidate selection while providing deterministic spatial tools, verification gates, multi-view renderers, and rubric aggregation.
Agent programs execute across seven isolated language runtimes; dashed circles indicate pluggable model interfaces.}
\label{fig:harness-architecture}
\end{figure}

\paragraph{Tracks and runtimes.}
\harness{} spans four generative tracks---static objects, articulated objects, 3D scenes, and procedural graphics---backed by seven isolated language runtimes: three for static objects (Blender Python, CadQuery, and Three.js), one each for articulated mechanisms (URDF with Blender-authored links) and scenes (Three.js), and two for graphics (GLSL and \texttt{moderngl}).
Each runtime specifies a strict file layout, static lint rules, a sandboxed build wrapper, canonical export formats, and a deterministic render driver, while sharing unified coordinate frames, metric units, camera rigs, and geometric tolerances.

\paragraph{Model roles and execution invariants.}
Each run decouples structural planning, code synthesis, and visual evaluation into three pluggable roles: a \emph{planner}, a \emph{coding agent}, and a \emph{vision--language judge} (Figure~\ref{fig:harness-architecture}).
The planner and judge operate through a unified API interface, while the coding agent executes inside an isolated workspace equipped with the harness's spatial tools via the Model Context Protocol (MCP), alongside a single-shot mode that requests a complete file envelope through direct API calls.
Every round commits raw source code from which meshes and renders are deterministically regenerated; boilerplate infrastructure (camera rigging, coordinate conversion, URDF composition, scene assembly, and shader headers) resides in the harness, and verification proceeds strictly from fast programmatic gates to blind VLM judging.

\subsection{Planning and the job graph}
\label{app:harness-plan}

\begin{figure}[!tbp]
\centering
\includegraphics[width=\linewidth]{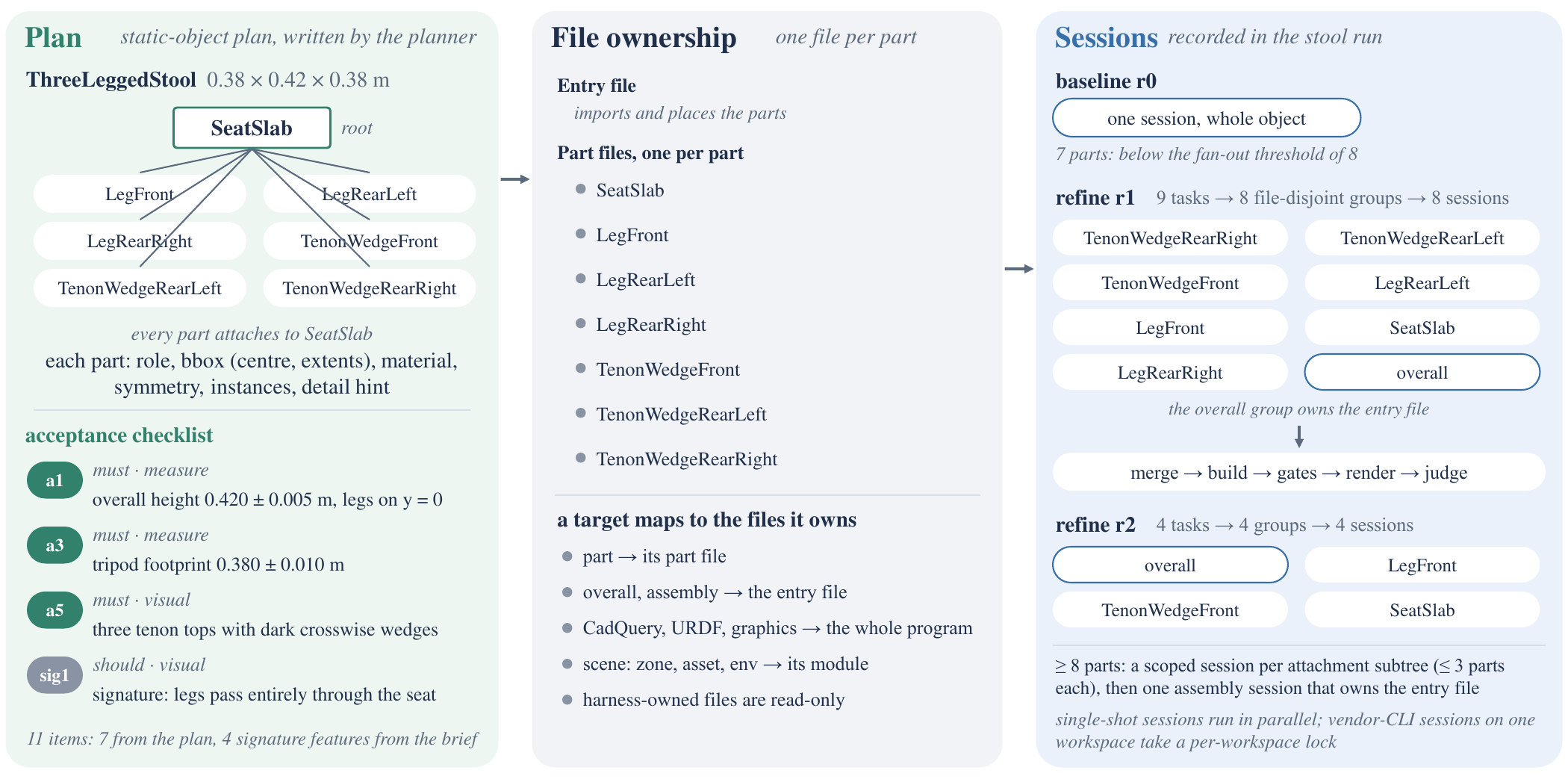}
\caption{\textbf{From structural plan to file scopes and coding sessions}, illustrated on the recorded trajectory of \S\ref{app:harness-example}.
The planner decomposes the prompt into named parts with metric bounding boxes, materials, attachment relations, and an acceptance checklist; each target maps to its owned source files, and refinement tasks sharing files are grouped into file-disjoint sessions.
In Round~1, nine revision tasks partition into eight file-disjoint groups; in Round~2, four tasks form four groups.}
\label{fig:harness-jobgraph}
\end{figure}

\paragraph{Structured specification and plan synthesis.}
The planner translates an open-ended request into a schema-validated structural plan: named parts with metric bounding boxes, materials, symmetry, and attachment topology for static objects; a kinematic tree with typed joints, axes, pivots, limits, and acyclic affine mimic couplings ($q_j = m_j q_i + b_j$) for articulated mechanisms; spatial zones, modular assets, effects, and authored cameras for scenes; and render passes, uniforms, and temporal dynamics for graphics.
Each plan carries an acceptance checklist binding items to an evidence modality (measurement, visual inspection, probe, or joint sweep) and a priority (\emph{must} or \emph{should}), supplemented by canonical rules that normalize kinematic conventions and enforce requested dimensions ($\pm 5\%$), triangle budgets, grounding, and temporal motion.

\paragraph{Bounded plan verification.}
Planning executes as a structured-output call with a token budget scaled to the target part count.
Range or formatting violations are repaired in-context, whereas a degenerate kinematic hierarchy referencing undeclared links triggers a clean context reset from the original request (\S\ref{app:loss-events}).

\paragraph{File-scoped subtree decomposition.}
Each runtime maps plan targets to strict file-ownership scopes (per-part modules plus an assembly entry file for Blender and Three.js objects; a single program for CadQuery, URDF, and graphics; and per-zone and per-asset modules for scenes), and writes outside a session's scope are reverted between Git commits.
Objects with eight or more top-level parts in modular languages are synthesized via scoped sessions over attachment subtrees of at most three parts---each supplied with the planned bounding boxes of adjacent parts---followed by an assembly session (Figure~\ref{fig:harness-jobgraph}).
During refinement, tasks that share files are grouped into connected components and dispatched as file-disjoint sessions, bounding per-session context and preventing cross-part regressions (\S\ref{app:loss-events}), while best-of-$N$ candidates and scene assets run concurrently in isolated workspaces.

\paragraph{Session context and contract injection.}
Every agent workspace is provisioned with a standardized instruction specification containing the runtime invariants, track-specific tool schemas, and language contracts.
Each task prompt supplies the user request, coordinate-frame conventions, the language cookbook, the structural plan formatted as exact metric tables with its acceptance checklist, and the verification protocol (requiring a clean build followed by spatial measurement or probing).
Single-shot prompts inline the code skeleton and request a complete file envelope in a single response.

\subsection{Spatial tools and deterministic gates}
\label{app:harness-tools}

\begin{figure}[!tbp]
\centering
\includegraphics[width=\linewidth]{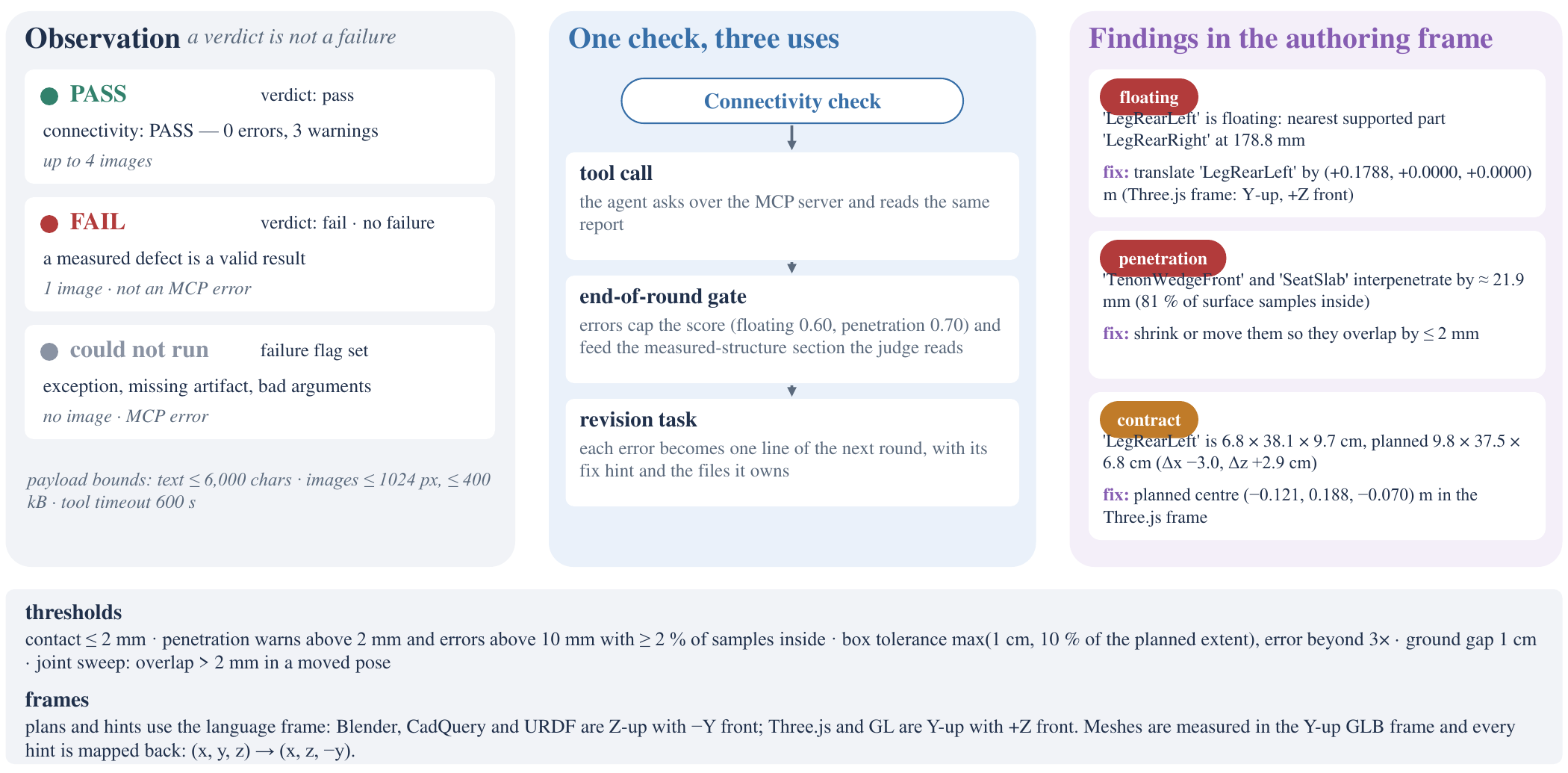}
\caption{\textbf{Observations, deterministic gates, and authoring-frame feedback.}
A measured geometric defect is returned as a structured negative verdict rather than a protocol error (left); a single verification routine serves simultaneously as an interactive agent tool, an end-of-round gate, and a generator of prioritized revision tasks (middle).
Right: baseline-round diagnostics from \S\ref{app:harness-example}, with corrective translation vectors projected directly into the agent's native authoring frame.}
\label{fig:harness-feedback}
\end{figure}

\paragraph{Semantic verdicts vs.\ transport errors.}
Every tool response separates the \emph{semantic verification verdict} (\textsc{pass}/\textsc{fail}) from the \emph{tool execution status}.
A detected mesh collision or build failure is returned as a valid observation leading with a structured \textsc{fail} verdict; only an internal tool crash sets the MCP protocol error flag, preventing commercial coding agents from entering degenerate transport-retry loops (\S\ref{app:loss-events}).
Tool responses bound text length and image resolution, attaching diagnostic renders specifically when a check identifies actionable geometric defects.

\paragraph{Deterministic gates, connectivity, and plan adherence.}
Following each successful build, the track pipeline executes its deterministic gates.
Gate errors compile into high-priority revision tasks with numerical fix hints and impose hard upper bounds on the round's rubric score, while warnings are surfaced to the judge.
Conversely, passing a deterministic gate vetoes contradictory visual defect claims from the VLM judge (\S\ref{app:harness-judge}).
In the connectivity gate, the ground-contacting component with the largest face count defines the primary support structure, and any disconnected component is reported alongside its nearest supported neighbour and the exact 3D translation vector that closes the gap.
A contact ledger tracks all planned part attachments with their measured clearances while enforcing structural support at the global assembly level.
The contract gate verifies built part counts, orientations, and bounding-box extents against the structural plan within per-axis tolerances.

\paragraph{Authoring-frame diagnostic projection.}
Although exported meshes are evaluated in the canonical Y-up frame, all corrective fix hints are projected back into each runtime's native authoring frame: a seat floating by $12$\,mm receives a corrective translation of $(0, 0, -0.012)$\,m in Z-up Blender Python and $(0, -0.012, 0)$\,m in Y-up Three.js (Figure~\ref{fig:harness-feedback}).
Articulated diagnostics output URDF joint origins and axes directly, and scene diagnostics emit native scene-graph API instructions.

\subsection{Automated judgment}
\label{app:harness-judge}

\begin{figure}[!tbp]
\centering
\includegraphics[width=\linewidth]{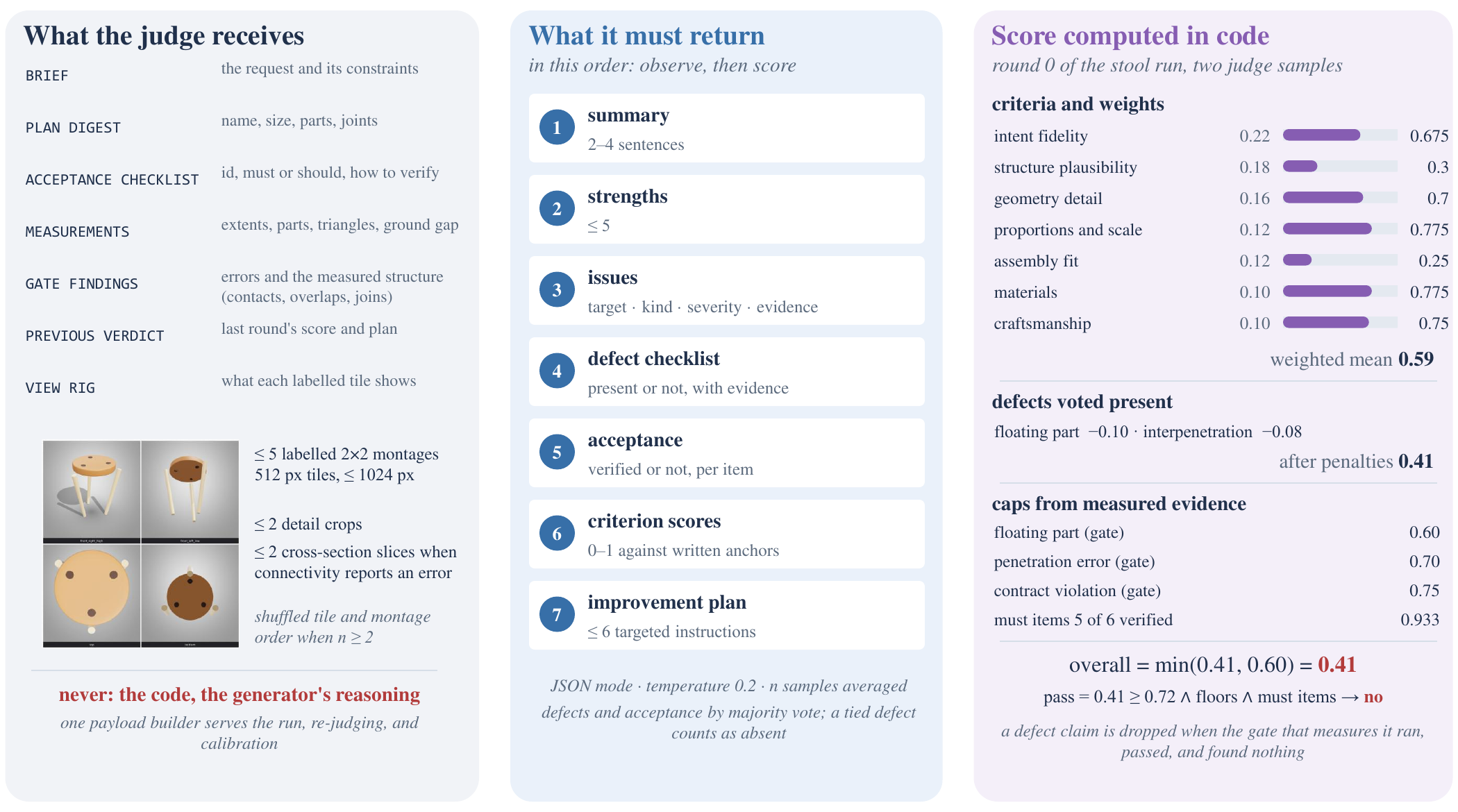}
\caption{\textbf{Vision--language judgment: evidence payload, structured schema, and score synthesis.}
The judge inspects structured telemetry and labelled multi-view montages without access to source code or generator reasoning (left); its schema enforces explicit visual observation prior to criterion scoring (middle).
Right: baseline-round score computation for \S\ref{app:harness-example}, combining criterion means, majority-voted defects, and deterministic gate caps.}
\label{fig:harness-judge}
\end{figure}

\paragraph{Evidence payload and response schema.}
A single payload builder serves in-loop refinement, post-run evaluation, and judge calibration (Figure~\ref{fig:harness-judge}).
The judge receives the request, plan digest, acceptance checklist, measurements, gate findings, and track telemetry, together with up to five labelled multi-view montages, detail crops, and---when connectivity errors occur---cross-sectional slices, with montage order randomized per sample to remove positional bias.
On object tracks, a passing connectivity gate prohibits unverified visual claims of floating or interpenetrating parts.
The judge answers in a strict JSON schema at temperature $0.2$, stating its evidence, strengths, issues, defect votes, and checklist verifications \emph{before} assigning criterion scores and up to six prioritized improvement instructions; deterministic criteria such as silhouette IoU are injected by code.

\paragraph{Score aggregation and deterministic vetoing.}
For round $t$, sample scores for each rubric criterion $k$ are averaged into $\bar{s}^{(t)}_k$, while checklist verifications and binary defects are aggregated by majority vote.
Any VLM-voted structural defect is automatically vetoed---incurring neither penalty nor cap---whenever the corresponding deterministic gate executed, passed without error, and recorded no penetration warning of $8$\,mm or deeper.
The composite round score is computed in code as
\begin{equation}
  s^{(t)} = \min\!\left\{u^{(t)},\,
    \max\!\left(0,\sum_k w_k\bar{s}_k^{(t)}-\sum_d\lambda_d\hat{d}^{(t)}_d\right)\right\},
  \label{eq:harness-score}
\end{equation}
where $\hat{d}^{(t)}_d\in\{0,1\}$ denotes retained defects with penalty weights $\lambda_d$, and $u^{(t)}\in[0,1]$ is the tightest upper bound among retained defect caps, triggered gate caps, and unverified mandatory checklist items ($u^{(t)}=1$ when none apply; on object tracks, verifying $v$ of $m$ mandatory items sets the checklist cap to $0.6 + 0.4\,v/m$).
A round achieves a \emph{pass} verdict when $s^{(t)}$ meets the track threshold, every criterion exceeds its floor, and all mandatory items are verified.
Every applied cap and its triggering evidence are archived in the verdict record.

\paragraph{Pairwise comparison and judge repeatability.}
For pairwise preference evaluation, the judge inspects anonymized candidates under both display orders and declares a winner only when both orders concur; reference-conditioned and likeness modes additionally incorporate measured silhouette IoU and real-world photographs.
Across repeated evaluations of fixed chair and cabinet artifacts under shuffled montage orders, Gemini~3.1~Pro achieves a mean score standard deviation of $0.030$ ($0.035$ under fixed order, compared with $0.083$ for Gemini~3.8~Flash), confirming that our primary evaluator provides a high-precision, view-order-robust scoring signal (\S\ref{app:eval-protocol}).

\subsection{Refinement rounds, history, and output selection}
\label{app:harness-rounds}

\paragraph{Round execution and task prioritization.}
Pre-round stages are cached, whereas every round executes fresh through a fixed verification chain: code synthesis, build with bounded repair, geometric measurement, deterministic gates, multi-view rendering, VLM judgment, and Git archival.
A failed build triggers up to three targeted repair sessions supplied with the error location, traceback, lint diagnostics, and relevant cookbook snippets, and every completed round is committed with a journal entry of its telemetry, verdicts, renders, and token usage.
Refinement tasks are compiled from the preceding round in strict priority order---deterministic gate errors with numerical fix hints, then unverified mandatory checklist items, then the judge's perceptual instructions---and deduplicated by part and defect category.
At most six tasks are dispatched per round, and truncation applies only to perceptual suggestions, so gate errors and mandatory items are never dropped; reference-conditioned runs append a silhouette task whenever outline IoU falls below $0.60$.

\paragraph{Termination and best-of-$N$ initialization.}
The standard profile allocates up to $T=4$ refinement rounds and a $60$-minute wall-clock ceiling; refinement stops early once all verification tasks are resolved, a round makes no code change, or the budget is reached, while monetary cost is logged for accounting only.
With multi-candidate initialization, the baseline round synthesizes $N$ independent candidates, ranks them lexicographically by build success, quick rubric score, and gate error count on a four-view rig, and seeds Round~$0$ with the winner.

\paragraph{Post-run output selection.}
Judged rounds are ranked by composite rubric score, breaking ties by fewer gate errors and earlier round index.
When the top two rounds fall within $0.03$, an optional pairwise VLM tie-breaker promotes the runner-up only if it wins both presentation orders with confidence $\ge 0.60$.
The selected round is packaged directly from its Git commit and verified artifacts without re-executing the build.

\subsection{A recorded run}
\label{app:harness-example}

\begin{figure}[!tbp]
\centering
\includegraphics[width=\linewidth]{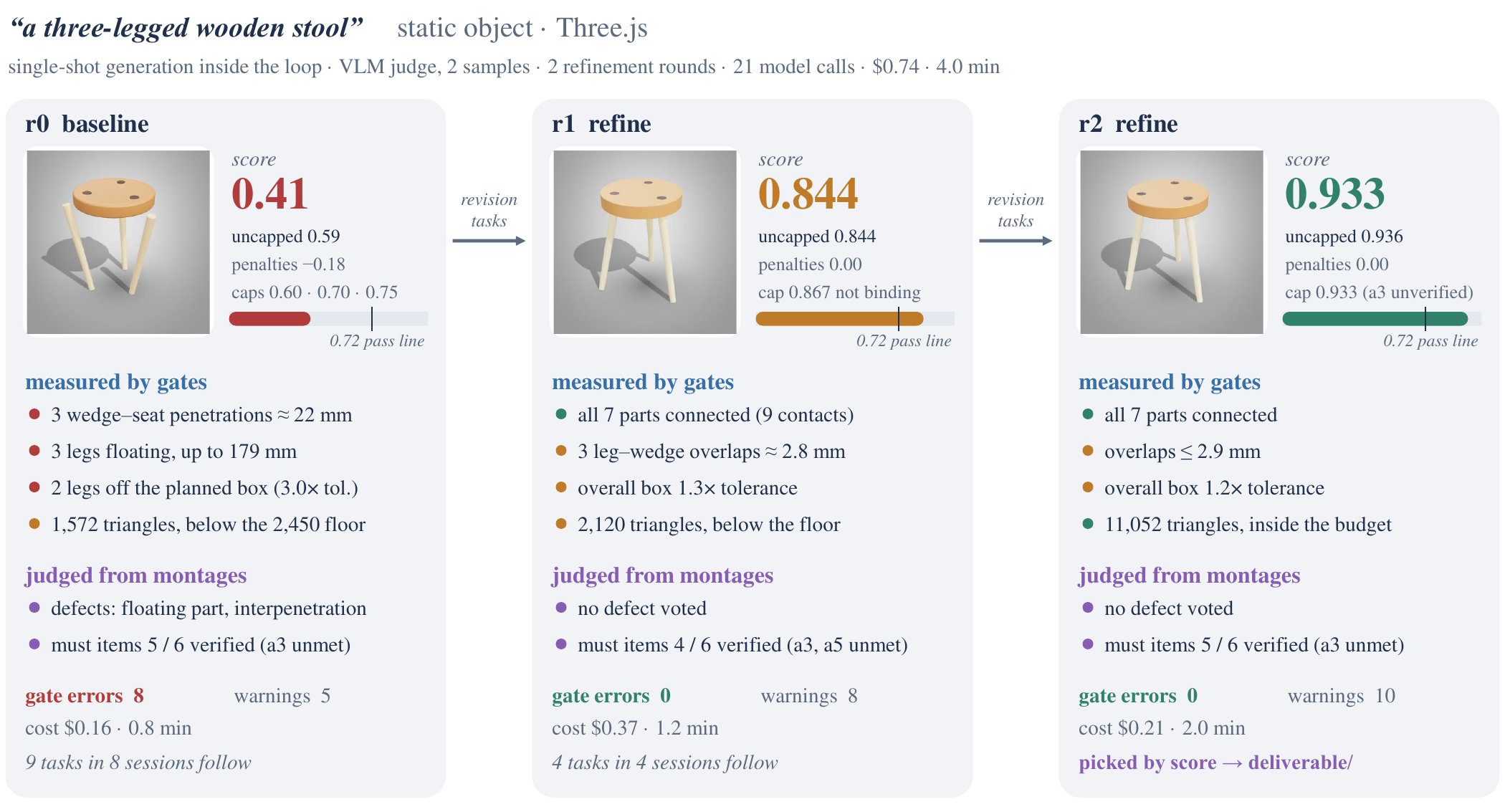}
\caption{\textbf{Recorded trajectory} for the prompt ``a three-legged wooden stool'' on the Three.js runtime under the economy profile.
Each column shows one round: the rendered view inspected by the judge, the composite score with its penalties and caps, deterministic gate findings, and VLM defect votes.
Closed-loop refinement resolves all baseline floating and interpenetrating parts in Round~1 ($0.410 \to 0.844$) and refines geometric detailing in Round~2 ($11{,}052$ triangles, reaching $0.933$ under the strict metric-footprint cap), completing in $4.0$ active minutes at \$0.74.}
\label{fig:refinement-example}
\end{figure}

Figure~\ref{fig:refinement-example} traces a complete run for the prompt ``a three-legged wooden stool'' on the Three.js runtime under the economy profile.
The planner decomposes the stool into seven parts (a circular seat slab, three splayed legs, and three through-tenon wedges) with eleven checklist items, including four signature structural features.
In the baseline round ($t=0$), the initial script builds, but deterministic gates identify multiple structural violations: the connectivity gate detects three wedge--seat interpenetrations ($\approx 22$\,mm) and six disconnected parts floating up to $212$\,mm from the support component; two legs exceed their planned bounding-box tolerances; and the mesh contains only $1{,}572$ triangles against a detail floor of $2{,}450$, yielding a composite score of $0.410$ ($0.59$ weighted criterion mean minus $0.18$ in defect penalties).
The harness compiles nine prioritized revision tasks---eight from deterministic gate errors with exact authoring-frame translation vectors and one from the footprint specification---and dispatches them across eight file-disjoint sessions.
By Round~1, every structural defect is resolved: all seven parts form a single ground-supported assembly with nine verified contacts, lifting the score by $+0.434$ to $0.844$.
Round~2 addresses the remaining geometric and metric refinements---carving recessed mortise slots, chamfering the slab rims, and increasing leg splay---expanding mesh detail to $11{,}052$ triangles and reaching a score of $0.933$ (bounded by the graded checklist cap of $0.6 + 0.4 \times 5/6$ for a $0.34$\,m vs.\ $0.38$\,m base diameter).
The complete trajectory finishes in $21$ model calls, $4.0$ active minutes, and \$0.74, and the output selector packages Round~2.

\subsection{Articulated objects: geometry, kinematics, and joint sweeps}
\label{app:harness-design}

\begin{figure}[!tbp]
\centering
\includegraphics[width=\linewidth]{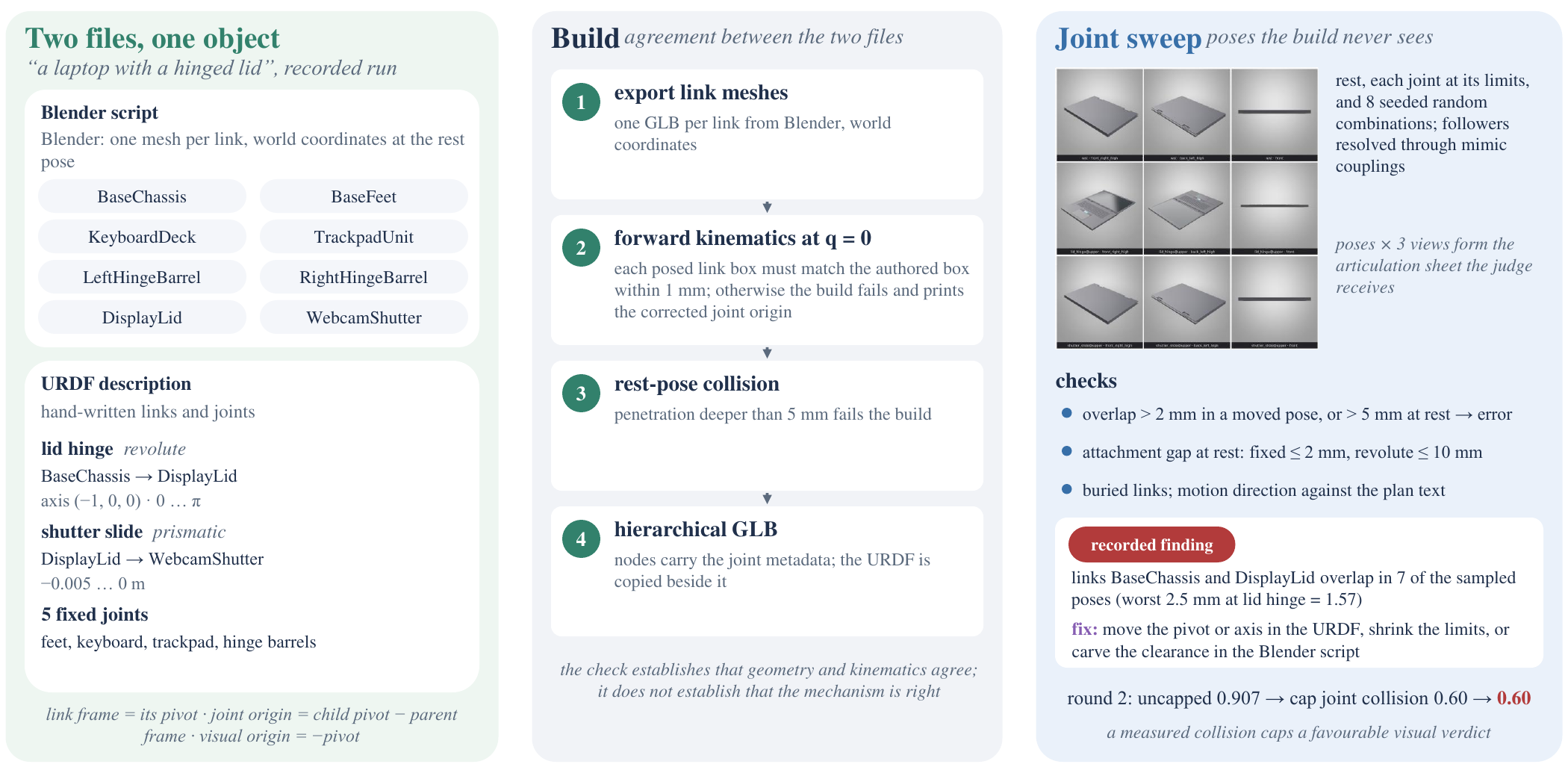}
\caption{\textbf{Articulated mechanism verification}, illustrated on a recorded run for ``a laptop with a hinged lid''.
Link geometry is authored in Blender and kinematic structure in URDF; the build verifies rest-pose frame agreement, and the joint sweep articulates the mechanism across its motion limits.
The dynamic sweep automatically detects a subtle $2.5$\,mm hinge--base collision at the upper opening limit that is invisible at rest, enforcing physical articulability via the rubric cap.}
\label{fig:crosslang}
\end{figure}

The articulated track coordinates two coupled source representations (Figure~\ref{fig:crosslang}): a Blender Python script authoring one mesh per link in world coordinates at the rest pose, and a URDF specification placing each link frame at its kinematic pivot.
Denoting the rest-pose frame of link $l$ by $\mathbf{T}_l(0)$ and its visual origin by $\mathbf{T}_{v,l}$, a world-authored link mesh is consistent with the kinematic tree if and only if
\begin{equation}
  \mathbf{T}_l(0)\,\mathbf{T}_{v,l}\,\mathbf{x}_{\mathrm{world}}
  = \mathbf{x}_{\mathrm{world}},
  \qquad
  \mathbf{T}_{v,l}=\mathbf{T}_l(0)^{-1}.
  \label{eq:urdf-frame}
\end{equation}
During the build stage, the harness compares each link's forward-kinematic bounding box at $q=0$ against its authored mesh; any alignment discrepancy exceeding $1$\,mm or rest-pose interpenetration exceeding $5$\,mm fails the build with an exact origin correction, guaranteeing rest-pose geometric--kinematic consistency prior to dynamic sweep evaluation.

The dynamic joint sweep then articulates the mechanism across its rest pose, the lower, midpoint, and upper limits of each independent joint, and eight seeded multi-joint configurations, resolving coupled followers analytically through the mimic graph.
Interpenetrations exceeding $2$\,mm in any articulated pose ($5$\,mm at rest) are flagged per link pair, while mobile child links separated by more than $30$\,mm from their parent at rest ($2$\,mm for fixed joints) are flagged as unattached.
A motion-direction check verifies that each joint's displacement vector aligns with the planned motion semantics and proposes axis corrections when inverted.
The VLM judge inspects the multi-pose articulation sheet alongside these geometric findings, while deterministic caps bound the score at $0.60$ for dynamic collisions or unattached links, $0.30$ for URDF schema errors, and $0.50$ for missing pose sheets (see Appendix~\ref{app:coupling-study} for kinematic coupling ablations).

\subsection{Scenes and graphics}
\label{app:harness-scene-graphics}

\begin{figure}[!tbp]
\centering
\includegraphics[width=\linewidth]{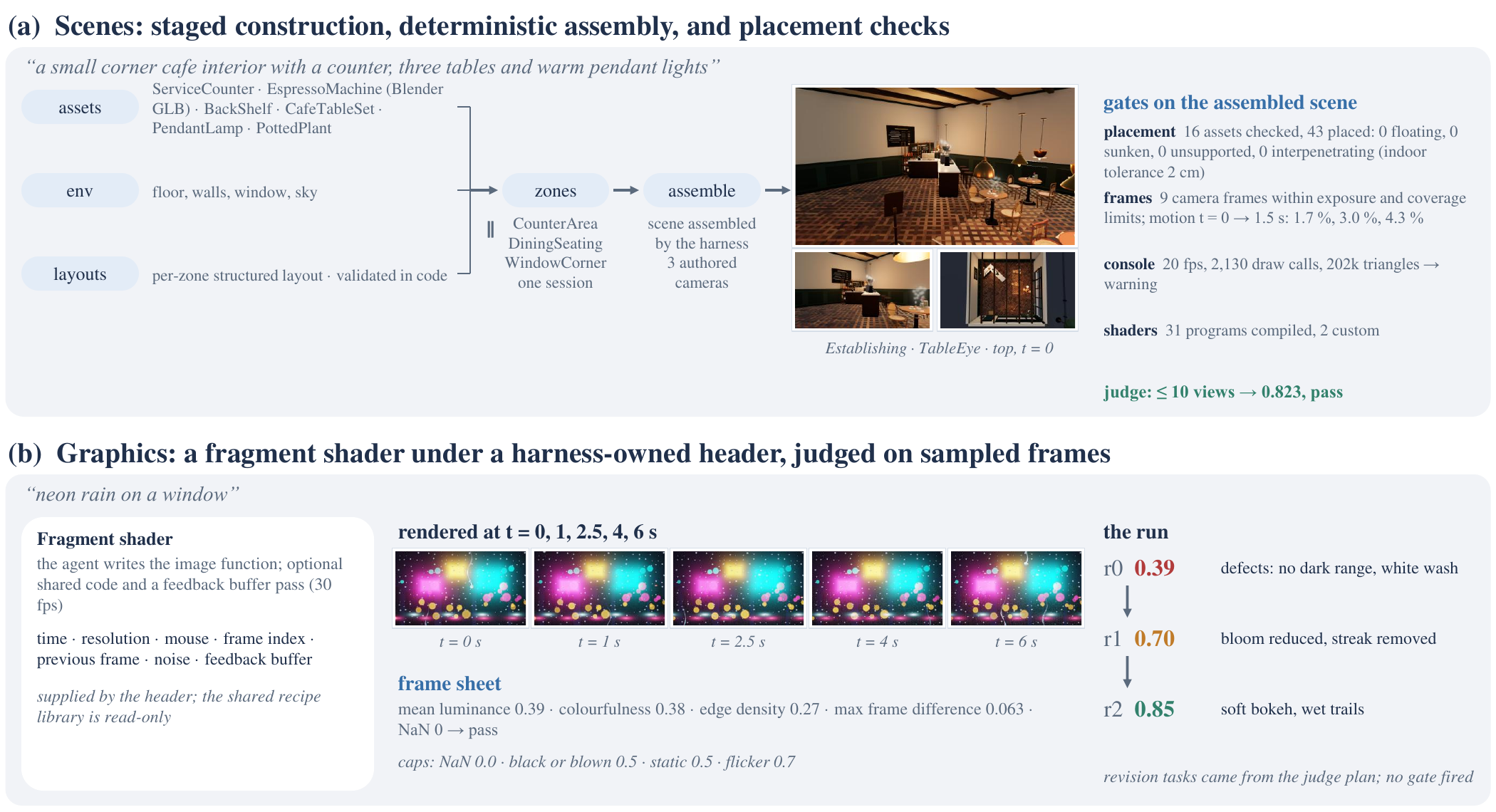}
\caption{\textbf{Scene and procedural graphics tracks.}
(a)~Scene generation executes a staged asset-and-zone assembly pipeline before entering the closed-loop refinement cycle with scene-specific gates.
(b)~Graphics generation synthesizes fragment shaders or OpenGL programs whose temporal frames are sampled at fixed timestamps, verified by statistical frame gates, and evaluated as multi-frame montages.}
\label{fig:scene-graphics}
\end{figure}

\paragraph{Scenes.}
Scene synthesis executes a staged hierarchical pipeline prior to the main refinement loop (Figure~\ref{fig:scene-graphics}a).
Individual scene assets (up to eight, or four under tight budget thresholds) follow a progressive compute-escalation policy: each asset begins with a fast single-shot draft and deterministic check, escalating to a full interactive agent session only if automated repair does not resolve all findings; Blender-built hero assets additionally undergo VLM evaluation with a monotonic quality gate that retains a revision pass only if it strictly improves the asset score.
Environmental terrains and spatial zone layouts are built concurrently, followed by a unified cross-zone composition session and deterministic scene-graph assembly.
To guarantee timely completion on complex prompts, pre-round staging is bounded at $55\%$ of the wall-clock budget, dynamically switching remaining pre-round stages to fast single-shot synthesis once reached.
Scene gates enforce asset grounding and collision bounds, runtime console cleanliness, per-camera exposure and framing coverage, inter-frame animation motion, asset utilization, and a $15{,}000$ draw-call budget across up to ten authored and overview camera views.

\paragraph{Graphics.}
On the graphics track, the agent authors either a Shadertoy-style GLSL fragment shader (with optional multipass feedback buffers) or a \texttt{moderngl} Python program defining setup and per-frame render routines, while the harness injects standardized headers, uniforms, outputs, and brief-matched reference recipes (Figure~\ref{fig:scene-graphics}b).
Sequences are rendered at $t\in\{0, 1, 2.5, 4, 6\}$\,s (clipped to the loop duration) at $1280\times720$ resolution.
The deterministic frame gate measures luminance distribution, colourfulness, edge density, inter-frame motion, and pixel validity: NaN, all-black, or overexposed frames trigger hard errors, while static or flickering sequences trigger score-capping warnings.
The judge evaluates multi-frame montages paired with frame telemetry, and each refinement round dispatches a unified whole-program optimization task.

\subsection{Skills}
\label{app:harness-skills}

\begin{table}[!tbp]
\centering
\caption{\textbf{The domain skill library}, listing each bundle's authoring guidance alongside its standing and gate-triggered routing conditions.}
\label{tab:skills}
\scriptsize
\setlength{\tabcolsep}{4pt}
\begin{tabularx}{\textwidth}{@{}l Y l l@{}}
\toprule
\textbf{Bundle} & \textbf{Guidance} & \textbf{Standing route} & \textbf{Gate-fired route} \\
\midrule
part-contact & contact, weld-overlap, penetration, and stray-island limits of the connectivity gate & multi-part objects & connectivity, joint sweep \\
bbox-contract & plan dimensions: tolerance $\max(1\,\mathrm{cm}, 10\%)$, error past $3\times$ & object tracks & contract \\
repeats-and-mirrors & instances, mirror and radial arrays, instance naming & instances or symmetry & instance box \\
blender-forms & form-to-Blender-API techniques; five common lint errors & Blender, URDF+Blender & part not imported \\
cadquery-forms & CadQuery forms; fillet, revolve, and boolean rules & CadQuery & --- \\
threejs-forms & form word to Three.js geometry; no CSG & Three.js objects & --- \\
urdf-joints & axes, limits, clearance; direction of $+q$ & articulated & joint sweep, motion direction \\
scene-composition & layout; cameras from measured bounds & scene sessions & camera, content, hero \\
scene-lighting & key, fill, and practical lights; linear-light albedo & scenes & dark, flat, blown frames \\
scene-motion & visible motion between the two animation times & scenes & no motion \\
threejs-shader-traps & custom GLSL that passes the shader preflight & custom shaders & shader preflight \\
glsl-craft & fragment-shader craft for the five sampled frames & GLSL graphics & frame gate \\
opengl-pipeline & moderngl setup and render contract; time sampling & OpenGL graphics & frame gate \\
scene-atmosphere & sky dome, environment bake, fog & scenes & --- \\
scene-water & water as a dielectric & water in plan & --- \\
scene-night & emissive windows, glow, light shafts & night in plan & --- \\
scene-materials & procedural albedo variance, per-instance jitter & scenes & --- \\
\bottomrule
\end{tabularx}
\end{table}

The authoring skill library (\S\ref{sec:corpus} and \S\ref{sec:harness}) comprises seventeen modular bundles in the open agent-skill format, each pairing a concise instruction sheet with a code reference file bounded at $350$ lines and ${\approx}2{,}500$ tokens (Table~\ref{tab:skills}).
A $28$-rule routing table dynamically selects up to five highest-priority bundles per session through two complementary mechanisms: \emph{standing routes}, which match the active track, runtime language, session role, and planned structural features, and \emph{diagnostic routes}, which trigger on specific gate findings from the preceding round (for example, attaching \texttt{threejs-shader-traps} whenever the shader preflight gate fires).

\FloatBarrier
\section{\corpus{}: Dataset Details}
\label{app:corpus-detail}

\subsection{Dataset hierarchy and counting units}
\label{app:corpus-snapshots}

\corpus{} is organized into three tiers that separate curated programs, benchmark training splits, and multi-representation exports.
The \emph{catalogue} indexes $269{,}597$ unique executable programs across nine collections (Table~\ref{tab:catalogue-legacy}), spanning parametric CAD, procedural Blender Python, OpenSCAD, articulated URDF mechanisms, Three.js scenes, and GLSL shaders.
The \emph{benchmark training pairs} ($768{,}596$ pairs, $891$\,M tokens) are deduplicated prompt--code splits, including web-animation programs, built for the fine-tuning experiments of \S\ref{sec:exp-sft} and Appendix~\ref{app:training}; they strictly exclude all \bench{} procedural factories, seeded instances, and per-dialect evaluation suites.
The \emph{unified text export} ($2{,}028{,}286$ rows across $342$ shards) expands each curated program across its three caption views (visual description, instruction, and factory recipe) and decomposes multi-pass shaders into pass-level targets: $891{,}788$ native-format rows, $848{,}819$ standardized text-export rows, and $287{,}679$ pass-flattened shader rows.

\begin{table}[t]
\centering
\scriptsize
\setlength{\tabcolsep}{3pt}
\caption{\textbf{The packaged catalogue}, one row per collection and dialect at the packaging snapshot.
Origin terms never replace the dataset-level CC-BY-NC-SA-4.0 terms; the stricter applicable terms control reuse.
\emph{Render entries} count the render paths recorded per sample, which include mesh exports next to images.
$^{\dagger}$Estimated from a $10\%$ stratified sample of archive manifests.
$^{\ddagger}$Indexed render paths in the initial release manifest; full eight-view and pose renders reside in the chunk archives (\S\ref{app:corpus-sources}).
$^{\S}$Packaged coding-agent subsets at the release snapshot (\S\ref{app:corpus-sources}).}
\label{tab:catalogue-legacy}
\begin{tabularx}{\textwidth}{@{}l l r r r r Y@{}}
\toprule
\textbf{Subset} & \textbf{Dialect} & \textbf{Samples} & \textbf{Share} &
\textbf{Code\,MB} & \textbf{Render entries} & \textbf{Origin terms} \\
\midrule
DeepCAD              & DeepCAD $\rightarrow$ CadQuery & 128{,}201 & 47.55\% & 136.7 &   769{,}107 & MIT \\
Shadertoy            & GLSL                &  119{,}961 & 44.50\% & 394.7 & 1{,}196{,}826$^{\dagger}$ & per shader, mostly CC BY-NC-SA \\
Articraft (CadQuery) & CadQuery            &    6{,}882 &  2.55\% & 120.0 &     3{,}300$^{\ddagger}$ & CC BY 4.0 \\
Bioinspired3D        & Blender Python      &    6{,}343 &  2.35\% &   9.3 &    31{,}715 & MIT \\
Articraft (URDF)     & URDF                &    6{,}146 &  2.28\% & 103.4 &     2{,}846$^{\ddagger}$ & CC BY 4.0 \\
Thingiverse-OpenSCAD & OpenSCAD            &    1{,}193 &  0.44\% &   4.9 &     5{,}965 & per item \\
\bench{} factories   & Blender Python      &       486 &  0.18\% &  16.0 &     2{,}430 & BSD-3-Clause \\
Three.js, agent-written$^{\S}$ & Three.js (JS/HTML) & 280 & 0.10\% & 3.4 & 1{,}120 & CC BY 4.0 per sample \\
Blender, agent-written$^{\S}$  & Blender Python & 80 & 0.03\% & 0.9 & 320 & CC BY 4.0 per sample \\
Three.js repositories & Three.js / R3F     &        25 &  0.01\% &  11.3 &         0 & as detected \\
\midrule
\textbf{Total}       &                     & \textbf{269{,}597} & \textbf{100.00\%} &
  \textbf{800.6} & \textbf{2{,}013{,}629} & dataset: CC-BY-NC-SA-4.0 \\
\bottomrule
\end{tabularx}
\end{table}

\subsection{Sources and conversion pipelines}
\label{app:corpus-sources}

\paragraph{DeepCAD to CadQuery.}
DeepCAD provides $178{,}238$ raw sketch--extrusion command sequences rather than executable programs~\citep{wu2021deepcad}.
Our deterministic compiler translates each command sequence into a self-contained CadQuery script: $174{,}981$ sequences compile ($98.2\%$), $174{,}177$ produce valid watertight solids ($97.7\%$), and vector-level hashing removes $25.8\%$ exact geometric duplicates to yield $128{,}201$ unique executable programs ($71.9\%$ of raw sequences).
Geometric fidelity against the original solids, measured via voxel IoU at $48^3$ resolution, averages $0.985$, with $96.9\%$ of programs exceeding $0.90$ (the remaining $1{,}114$ programs are annotated with explicit fidelity metadata so downstream pipelines can apply custom thresholds).
Natural-language annotations from Text2CAD~\citep{khan2024text2cad} are paired with $125{,}162$ programs under their original license terms, accompanied by four canonical views per solid.
Crucially, because the upstream DeepCAD split contains $3{,}230$ duplicate geometry groups straddling its train and test partitions, our vector-and-mesh deduplication eliminates cross-split leakage across the CAD lineage.

\paragraph{Shadertoy.}
The procedural graphics collection comprises $119{,}961$ Shadertoy shaders---$98{,}164$ single-pass and $21{,}585$ multi-pass programs---indexed with one canonical record per shader~\citep{shadertoy}.
Each record stores the primary image pass as the indexed entry file alongside the complete multi-pass buffer source tree, multi-frame render grids, video clips, and bound input textures.
For text-to-code training, multi-pass shaders are additionally expanded by prepending shared common headers to each buffer pass ($287{,}679$ pass-level rows); automated syntax and uniform repair across $588$ shaders lifts the headless compilation rate of the flattened set from $84.6\%$ to \textbf{99.0\%} ($92{,}534$ of $93{,}434$), with only compiled shaders admitted to training mixtures.

\paragraph{Articraft.}
The $9{,}996$ upstream Articraft records~\citep{zhou2026articraft} partition into $6{,}882$ mesh-based CAD assemblies and $3{,}114$ primitive-based articulated mechanisms.
For the mesh-based assemblies, we perform a clean-room reimplementation of the Articraft geometry helpers coupled with a tree-shaking AST inliner to produce standalone CadQuery programs (median ${\approx}900$ lines), verifying every output against the upstream geometry via Chamfer distance ($< 0.02$) and bounding-box aspect ratio ($\in [0.9, 1.1]$); $6{,}878$ of $6{,}882$ ($99.9\%$) pass verification and are rendered from eight canonical views.
The primitive-based mechanisms are packaged as executable URDF specifications across geometry-only and textured variants ($6{,}146$ records in total, including $3{,}032$ textured mechanisms) with multi-pose kinematic renders.

\paragraph{Procedural, community, and agent-synthesized collections.}
The remaining sources (Table~\ref{tab:catalogue-legacy}) are all execution-verified:
(1)~\emph{Bioinspired3D}~\citep{bioinspired3d}, $6{,}343$ Blender Python programs ($4{,}557$ geometry programs and $1{,}786$ material-textured variants) that all execute, export, and render from five views;
(2)~\emph{Thingiverse OpenSCAD}, $1{,}193$ parametric projects curated from $7{,}378$ community designs, with their original source files, five renders, and item-level license metadata;
(3)~\emph{\bench{} factories}, $486$ procedural Blender generators ($243$ factories derived from Infinigen~\citep{raistrick2023infinigen}, each in geometry and textured variants, averaging ${\approx}880$ lines) and $3{,}904$ curated seeded instances ($1{,}953$ geometry and $1{,}951$ textured), all held out from benchmark training because the $212$ \bench{} tasks come from the same generator family (Appendix~\ref{app:training});
(4)~\emph{agent-authored programs}, $949$ Blender Python programs and $1{,}134$ Three.js scenes synthesized by coding agents under execution and render verification (from $1{,}040$ and $1{,}389$ attempts), plus $25$ multi-file Three.js repositories and $1{,}051$ web-animation programs across $21$ categories; and
(5)~\emph{harness-generated bootstrap programs}, $15{,}704$ execution-verified programs carrying quality tiers, checklist verdicts, and gate telemetry, which substantially improve target-dialect executability in our fine-tuning mixtures (\S\ref{app:model-baselines}).

\FloatBarrier
\section{Experimental Details and Additional Results}
\label{app:experiments}

\subsection{Training details}
\label{app:training}

\paragraph{Training configuration.}
All fine-tuning runs are conducted with LLaMA-Factory~\citep{zheng2024llamafactory} on $4\times\text{NVIDIA H100}$ ($80$\,GB) GPUs in \texttt{bfloat16} precision using gradient checkpointing and sequence packing.
Supervised fine-tuning (SFT) applies LoRA~\citep{hu2022lora} of rank $64$ ($\alpha=128$, dropout $0.05$) across all linear projections with a learning rate of $1\times 10^{-4}$ (or $1\times 10^{-5}$ for full-parameter fine-tuning under ZeRO-3~\citep{rajbhandari2020zero}), a cosine learning-rate schedule, a $5\%$ linear warm-up ratio, and a per-device batch size of $1$ with gradient accumulation of $8$ ($4$ for $27$B).
Direct Preference Optimization (DPO;~\citealt{rafailov2023dpo}) initializes from the merged SFT checkpoint and trains a rank-$16$ LoRA ($\alpha=32$) for $2$ epochs under sigmoid preference loss ($\beta=0.1$, learning rate $5\times 10^{-6}$).
The sequence context length is set to $8{,}192$ tokens for $9$B runs and $4{,}096$ tokens for $27$B runs to fit single-device memory bounds under data-parallel training.
Evaluation inference uses vLLM~\citep{kwon2023vllm} under greedy decoding or temperature sampling ($T{=}0.7$) as specified per experiment.

\paragraph{Training throughput.}
Because LoRA introduces a compact optimizer state, training the $27$B model under standard Distributed Data Parallel (DDP) avoids ZeRO-3 parameter all-gather traffic and achieves a \textbf{13$\times$ higher step throughput} (${\approx}6.2$\,K vs.\ ${\approx}460$ tokens/s) with standard PyTorch allocator caching.
Across $4\times\text{H100}$ GPUs, $9$B LoRA sustains ${\approx}17$\,K tokens/s (completing the $270$\,K-pair, $293$\,M-token mixture in $4$\,h\,$46$\,m), while $27$B LoRA sustains ${\approx}6.2$\,K tokens/s.

\paragraph{Chat formatting and strict contamination control.}
Training examples are formatted as ShareGPT multi-turn conversations: a system prompt specifying the target dialect and output contract, a user turn containing the natural-language specification, and an assistant turn containing the complete executable program in a fenced code block.
Any example whose tokenized conversation exceeds the context window is excluded cleanly rather than truncated, preserving syntactic closure.
To guarantee zero train--test leakage, all $212$ \bench{} procedural factories, all $3{,}904$ re-seeded instances derived from those factories, and all per-dialect evaluation splits are strictly held out; no training mixture shares a program or asset identifier with any evaluation task.
Table~\ref{tab:mixes} details the exact dialect composition, token volume, and wall-clock training duration for every mixture.

\begin{table}[ht]
\centering
\caption{\textbf{Supervised fine-tuning mixtures.}
Composition, token volume, and wall-clock training duration on $4\times\text{H100}$ GPUs across the SFT mixtures evaluated in \S\ref{sec:experiments} and \S\ref{app:model-baselines} (\emph{boot} denotes execution-verified bootstrap programs synthesized by \harness{}; \emph{cap/sample} denotes the number of caption views per program).}
\label{tab:mixes}
\small
\setlength{\tabcolsep}{4pt}
\resizebox{\textwidth}{!}{%
\begin{tabular}{@{}l l r r c l r@{}}
\toprule
\textbf{Mix} & \textbf{Base} & \textbf{Pairs} & \textbf{Mtok} &
\textbf{cap/sample} & \textbf{Composition} & \textbf{Wall} \\
\midrule
5.2\,K Blender & 9B & 5{,}193 & 3.8 & mixed & Bioinspired3D + non-test factories & 30\,m \\
25.6\,K        & 9B & 25{,}600 & 26 & 1 & 4 dialects, merged single-dialect sets & 1\,h46 \\
102\,K         & 9B & 101{,}851 & 107 & 1 & CadQ.\ 40\,K, GLSL 40\,K, OpenSCAD all, Blender 5\,K + 15.7\,K boot & 3\,h25 \\
116\,K \emph{(bal)} & 9B & 116{,}000 & 110 & 1 & as above, Blender share 15\%\,$\to$\,34\% by \emph{repetition} & 2\,h26 \\
\textbf{270\,K} & 9B & 270{,}173 & 293 & 1 & all CadQuery, all GLSL, Blender 5\,K + 15.7\,K boot, OpenSCAD $\times$4 & 4\,h46 \\
364\,K         & 9B & 364{,}000 & 304 & \textbf{3} & all six sources, \textbf{no boot} & 6\,h43 \\
273\,K         & 9B & 273{,}000 & 300 & 1 & all six sources, \textbf{+ all boot} & 4\,h50 \\
\midrule
76\,K          & 27B & 75{,}910 & 56 & $\leq 2$ & CadQuery $+$ GLSL = 74\% of tokens, no boot & 3\,h51 \\
\textbf{104\,K}& 27B & 103{,}770 & 75 & $\leq 2$ & Blender share \textbf{40\%}, all boot restored & 3\,h35 \\
\bottomrule
\end{tabular}}
\end{table}

\subsection{Training curves of the 27B data-scale runs}
\label{app:training-loss}

This section characterizes the optimization dynamics of Qwen3.8-27B across the representation tiers of \corpus{}.
Three LoRA runs share an identical optimization recipe and vary solely in which representation views of the unified text export (Appendix~\ref{app:corpus-snapshots}) they consume.

\paragraph{Optimization recipe.}
Each run applies LoRA of rank $16$ ($\alpha=32$, dropout $0.05$) for one epoch at a peak learning rate of $5\times 10^{-5}$ with a linear warm-up over the first $3\%$ of steps and cosine decay to zero.
Examples are packed into full sequences (${\approx}7.2$ samples per sequence), with each optimizer step consuming $64$ packed sequences and logging training cross-entropy every five steps.

\paragraph{Representation scaling arms.}
\textbf{Arm~A} trains on the complete unified export across all three representation views (native format, standardized text export, and pass-flattened multi-pass shaders): $1{,}992{,}816$ rows packed into $276{,}594$ sequences ($4{,}322$ steps).
\textbf{Arm~B} trains on the native-format and pass-flattened views without the re-templated text-export variant: $1{,}160{,}630$ rows in $159{,}427$ sequences ($2{,}492$ steps).
\textbf{Arm~C} retains a single canonical copy of each curated program paired with its instruction-style caption: $292{,}690$ rows in $39{,}926$ sequences ($647$ steps).

\begin{figure}[!tbp]
\centering
\includegraphics[width=\linewidth]{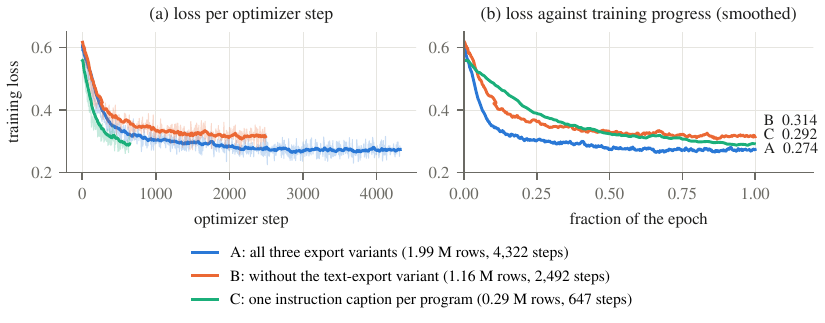}
\caption{\textbf{Training loss dynamics of the three Qwen3.8-27B data-scale runs.}
\textbf{(a)}~Cross-entropy loss per optimizer step; faint curves show raw 5-step logs and solid curves show an exponential moving average ($\tau = 50$ steps).
\textbf{(b)}~Smoothed loss trajectories plotted against epoch fraction to align cosine learning-rate schedules, annotated with terminal smoothed losses.
Each trajectory tracks in-distribution optimization across its respective representation mixture from $293$\,K to $1.99$\,M rows.}
\label{fig:training-loss}
\end{figure}

\paragraph{Optimization dynamics across representation scales.}
Figure~\ref{fig:training-loss} demonstrates smooth, stable convergence across all three regimes: no logged step exceeds $1.3\times$ the trailing $20$-step moving average, and all runs are entirely free of instability or divergence spikes.
Rapid initial adaptation occurs within the first quarter of the epoch---reducing cross-entropy from ${\approx}0.60$ to $0.30$ (Arm~A) and $0.36$ (Arms~B and~C)---followed by steady refinement to terminal smoothed losses of $0.274$ (Arm~A), $0.314$ (Arm~B), and $0.292$ (Arm~C), with Arm~A stabilizing across the second half of the epoch ($0.277$ at mid-epoch to $0.274$ at completion).
Together, the three arms highlight two complementary operating regimes of \corpus{}: Arm~C ($293$\,K unique instruction--program pairs, $647$ steps) offers high sample efficiency for rapid adaptation, achieving a lower terminal loss than Arm~B in one quarter of the optimization steps, whereas Arm~A ($1.99$\,M rows, $4{,}322$ steps) leverages multi-caption and pass-level shader views to reinforce cross-representation alignment, achieving the lowest loss both at matched step budgets ($0.275$ vs.\ $0.316$ at step $2{,}490$) and at full convergence ($0.274$).

\subsection{Evaluation protocol}
\label{app:eval-protocol}

\paragraph{Code extraction and geometric evaluation for open models.}
For open-model evaluations (\S\ref{sec:exp-sft}), generated responses pass through a deterministic code extractor that strips reasoning blocks, collects fenced code blocks (including trailing unterminated blocks), and ranks candidates by dialect markers, fence annotations, and AST syntactic validity.
This extraction step is essential for pretrained reasoning checkpoints: with thinking enabled, $52$ of $212$ responses from the zero-shot $27$B model omit markdown fences and $74$ place auxiliary snippets after the main program, whereas fine-tuned models adhere strictly to the single-block output contract ($100\%$ single-fence compliance).
Extracted programs execute inside the dialect's pinned runtime under a $300$\,s timeout.
Resulting meshes are exported to GLB, centred at the origin, normalized to unit maximum extent, uniformly sampled at $10{,}000$ surface points, and evaluated via symmetric Chamfer-$L_2$ distance and F-score at thresholds $\tau \in \{0.05, 0.10\}$ (taking the best alignment across the four $90^\circ$ yaw rotations; ground-truth reference scripts self-verify at $\text{Chamfer} \approx 0.016$ and $\text{F@0.05} \approx 1.0$).

\paragraph{Object prompts and compared systems.}
\label{app:harness-object-protocol}
The 3D object benchmark comprises $100$ Three.js generation prompts stratified across three specification levels---$21$ concise (\emph{simple}), $17$ moderate (\emph{medium}), and $62$ richly specified (\emph{detailed}) prompts---testing synthesis across varying levels of structural and stylistic complexity without prescribing explicit metric dimensions or part lists.
All systems are evaluated under a uniform \emph{single-request deliverable protocol}: each workflow receives the identical user prompt string and delivers a final executable program, which is rebuilt and scored by our common evaluator.
For commercial product baselines such as Claude Design (powered by Claude Opus~5.5), we evaluate its released reference outputs directly in our unified evaluation pipeline without modification.

We evaluate Gemini~3.8~Flash and Claude Opus~5.5 against their respective native baselines on the primary $60$-prompt benchmark, and evaluate Claude Sonnet~5, GPT-6~Luna, and GPT-6~Astra on a balanced $30$-prompt subset ($10$ prompts per specification level).
Each native baseline (Table~\ref{tab:harness}) synthesizes code in a single pass using its standard coding-agent backend (high reasoning effort for Claude Opus~5.5 and Sonnet~5; medium effort for GPT-6~Astra; direct API for Gemini~3.8~Flash), whereas \harness{} wraps the identical backbone model with structured planning, spatial tools, and automated verification.
This paired design directly isolates the end-to-end gain of the \harness{} workflow for a fixed backbone model, with token telemetry and API-equivalent cost reported alongside quality metrics (Table~\ref{tab:harness-object-cost}).

\paragraph{Single-request autonomy and refinement horizons.}
The single-request protocol prohibits human-in-the-loop intervention: each system receives one user prompt with zero subsequent human edits or manual candidate cherry-picking, while permitting autonomous internal tool calls and verification passes.
Our primary object benchmark evaluates single-round harness execution ($T=0$ in the notation of \S\ref{sec:harness}), retaining the initial planned generation, interactive tool use, automated build repair, and final verification (using Gemini~3.8~Flash as planner and Gemini~3.1~Pro as in-loop judge for the Flash runs).
Multi-round studies ($T=4$) extend this autonomous loop with up to four closed-loop refinement rounds and automated post-run output selection under the same single user request.

\paragraph{Independent evaluation judge.}
\label{app:eval-judge}
All Judge Scores and VLM Preferences reported in \S\ref{sec:experiments} are produced by a single frozen evaluation pipeline that applies the judgment architecture and track rubrics of Appendix~\ref{app:harness-judge} independently of the generation workflow under test.
Each delivered program is rebuilt from scratch in its pinned runtime, with failed builds scored zero, to extract geometric measurements, connectivity or kinematic-sweep telemetry, and canonical multi-view renders (fourteen shaded views plus clay views for objects, multi-pose sweeps for articulated mechanisms, authored and orbital views at two timestamps for scenes, and sampled frame sheets for graphics); Gemini~3.1~Pro then scores it three times.
Three safeguards decouple evaluation from in-loop refinement:
(1)~\emph{clean-room rebuild and blind prompting}, scoring every artifact against the raw user prompt and a prompt-derived checklist alone, with all harness-generated plans, internal checklists, and refinement histories stripped;
(2)~\emph{deterministic geometric caps}, bounding scores via exact mesh connectivity and kinematic checks independent of VLM perception (Eq.~\ref{eq:harness-score}); and
(3)~\emph{multi-metric validation}, pairing Judge Scores with SigLIP-2 similarity and blind human pairwise preferences (\S\ref{sec:experiments}), which consistently corroborate the rubric gains across all four tracks.

\paragraph{Visual-criterion ablation.}
Our primary metric, the overall Judge Score $S_{\mathrm{all}}$ (Table~\ref{tab:harness}), combines seven weighted object criteria---intent fidelity ($0.22$), structural plausibility ($0.18$), geometry detail ($0.16$), proportions and scale ($0.12$), assembly fit ($0.12$), materials ($0.10$), and craftsmanship ($0.10$)---with deterministic defect penalties and caps (Eq.~\ref{eq:harness-score}).
To verify that our gains over external baselines such as Claude Design do not stem merely from metric-scale conventions or hard structural caps on open-ended prompts, we additionally compute an uncapped, scale-free visual score in Tables~\ref{tab:harness-object-scores} and~\ref{tab:harness-track-scores}:
\begin{equation}
S_{\mathrm{vis}}=\frac{\sum_{k\ne\mathrm{scale}} w_k s_k}{\sum_{k\ne\mathrm{scale}}w_k},
\label{eq:harness-visual-score}
\end{equation}
which aggregates raw criterion means $s_k$ prior to aggregate defect penalties and score caps while excluding the proportions-and-scale criterion (with failed builds scored zero on both metrics).
Even under this forgiving, uncapped metric, \harness{} outperforms Claude Design by $+25.6$ points ($88.1$ vs.\ $62.4$, winning $58$ of $60$ prompts), confirming that its advantage reflects superior geometric fidelity and visual craftsmanship rather than rubric thresholding.

\paragraph{SigLIP-2 text--render similarity.}
We compute SigLIP-2~\citep{tschannen2025siglip2} (\texttt{So400m}, patch size $16$, NaFlex) cosine similarity at a frozen checkpoint revision between the raw user prompt and four canonical shaded views of each object (elevated front-right, back-right, back-left, and front-left views, excluding contact sheets and clay renders).
Using the unmodified prompt without classification templates, we $\ell_2$-normalize text and image embeddings and average raw cosine similarities across the four views:
\begin{equation}
S_{\mathrm{SigLIP}}(c,f)=\frac{1}{4}\sum_{v=1}^{4}
\frac{E_{\mathrm{text}}(c)^\top E_{\mathrm{image}}(I_v(f))}
{\|E_{\mathrm{text}}(c)\|_2\,\|E_{\mathrm{image}}(I_v(f))\|_2}.
\end{equation}
Table~\ref{tab:harness} averages these object-level similarities over the $60$-prompt benchmark, assigning zero to failed builds.

\paragraph{Pairwise preference protocol.}
Both human and VLM pairwise evaluations present the original user prompt alongside anonymized multi-view renders captured under matched camera poses, framing, and resolution.
Evaluators judge overall preference on prompt adherence, structural plausibility, and visual craftsmanship, with ties counted as half a win; VLM preference additionally requires agreement across both presentation orders (\S\ref{app:harness-judge}).
Human evaluations randomize left/right presentation order across a fixed, pre-registered prompt subset shared by all compared systems, with prompt-level bootstrap confidence intervals clustering repeated ratings at the task level.
In end-to-end preference reporting, a valid executable artifact wins against a failed build and two failed builds tie, while conditional preference over mutually executing pairs is tracked alongside execution rate; transient provider outages are excluded from win/loss tallies.

\paragraph{Object and cross-track score breakdown.}
\label{app:harness-current}
Tables~\ref{tab:harness-object-scores}--\ref{tab:harness-track-scores} report paired score comparisons and $95\%$ Student-$t$ confidence intervals across objects, scenes, and procedural graphics (where W/L/T tallies the signs of per-prompt score differences).
On the $60$-prompt object benchmark---which concentrates heavily on complex specifications ($52$ detailed, $4$ medium, and $4$ simple prompts)---\harness{} lifts Gemini~3.8~Flash by $+25.3$ points ($86.6$ vs.\ $61.3$, winning $56$ of $60$ prompts) and outperforms Claude Design by $+49.2$ points ($59$--$1$ W/L/T), even surpassing the native Claude Opus~5.5 baseline by $+9.4$ points ($86.6$ vs.\ $77.2$).
These gains generalize across tracks (Table~\ref{tab:harness-track-scores}): \harness{} yields large improvements on procedural graphics for GPT-6~Astra ($+30.4$ points, $10$--$1$ W/L/T), Gemini~3.8~Flash ($+27.2$, $10$--$1$), and GPT-6~Luna ($+6.8$, $8$--$3$), as well as on 3D scenes for Flash ($+19.4$, $9$--$1$) and Luna ($+14.6$, $7$--$0$).
For Claude Sonnet~5 under single-round ($T=0$) execution, \harness{} wins the majority of graphics prompts ($6$--$5$ W/L/T) and improves uncapped scene quality ($\Delta S_{\mathrm{vis}} = +2.8$), while overall mean scores reflect strict deterministic gate caps applied to outlier frames prior to multi-round refinement.

\begin{table}[!ht]
\centering
\caption{\textbf{Paired object score comparisons ($0$--$100$ scale).}
Upper panel: composite Judge Score $S_{\mathrm{all}}$ incorporating deterministic defect penalties and caps.
Lower panel: uncapped, scale-free visual score $S_{\mathrm{vis}}$ (Eq.~\ref{eq:harness-visual-score}).
W/L/T tallies per-prompt wins, losses, and ties by score difference.}
\label{tab:harness-object-scores}
\footnotesize
\setlength{\tabcolsep}{3pt}

\begin{tabular*}{\linewidth}{@{\extracolsep{\fill}}l r r r r c c@{}}
\toprule
\textbf{A vs.\ B} & $n$ & $S_{\mathrm{all}}^{A}$ & $S_{\mathrm{all}}^{B}$ & $\Delta S_{\mathrm{all}}$ & \textbf{95\% CI} & \textbf{W/L/T} \\
\midrule
Flash + \ourmethod{} vs.\ Flash baseline & 60 & 86.6 & 61.3 & $+25.3$ & $[19.0,31.6]$ & 56/4/0 \\
Flash + \ourmethod{} vs.\ Claude Design & 60 & 86.6 & 37.4 & $+49.2$ & $[43.6,54.9]$ & 59/1/0 \\
Opus baseline vs.\ Claude Design & 60 & 77.2 & 37.4 & $+39.9$ & $[34.5,45.2]$ & 59/1/0 \\
Flash baseline vs.\ Claude Design & 60 & 61.3 & 37.4 & $+23.9$ & $[17.3,30.5]$ & 51/9/0 \\
\midrule
\textbf{A vs.\ B} & $n$ & $S_{\mathrm{vis}}^{A}$ & $S_{\mathrm{vis}}^{B}$ & $\Delta S_{\mathrm{vis}}$ & \textbf{95\% CI} & \textbf{W/L/T}$^{*}$ \\
\midrule
Flash + \ourmethod{} vs.\ Flash baseline & 60 & 88.1 & 69.3 & $+18.8$ & $[13.2,24.5]$ & 54/6/0 \\
Flash + \ourmethod{} vs.\ Claude Design & 60 & 88.1 & 62.4 & $+25.6$ & $[22.2,29.1]$ & 58/2/0 \\
Opus baseline vs.\ Claude Design & 60 & 84.5 & 62.4 & $+22.0$ & $[19.1,25.0]$ & 60/0/0 \\
Flash baseline vs.\ Claude Design & 60 & 69.3 & 62.4 & $+6.8$ & $[1.4,12.2]$ & 45/15/0 \\
\bottomrule
\end{tabular*}
\par\raggedright
$^{*}$Lower-panel W/L/T uses $S_{\mathrm{vis}}$ differences.
Flash denotes Gemini~3.8~Flash and Opus denotes Claude Opus~5.5; baselines synthesize code in a single pass without spatial tools or verification feedback (Appendix~\ref{app:harness-object-protocol}).

\end{table}

\begin{table}[!ht]
\centering
\caption{\textbf{Object execution reliability, token telemetry, and prompt-cache efficiency.}
Usage is averaged across all $60$ prompts and excludes post-run evaluation calls.
Over $93.9\%$ of \harness{} input tokens are served from low-cost prompt cache.}
\label{tab:harness-object-cost}
\footnotesize
\setlength{\tabcolsep}{3pt}

\begin{tabular*}{\linewidth}{@{\extracolsep{\fill}}l r r r r r r@{}}
\toprule
\textbf{System} & $n$ & \textbf{Exec.\ (\%)} & \textbf{Input (k)} & \textbf{Cached (k)} & \textbf{Output (k)} & \textbf{Est.\ USD} \\
\midrule
Claude Design & 60 & 100.0 & -- & -- & -- & -- \\
Flash baseline & 60 & 93.3 & 0.3 & 0.0 & 9.4 & 0.035 \\
Opus baseline & 60 & 100.0 & 3.0 & 0.5 & 28.5 & 0.353 \\
Flash + \ourmethod{} & 60 & \textbf{100.0} & 20{,}816.0 & 19{,}551.8 & 234.9 & 3.312 \\
\bottomrule
\end{tabular*}
\par\raggedright
Cached tokens are a subset of input tokens; output tokens include reasoning traces.
Claude Design artifacts are evaluated directly from released reference programs.

\end{table}

\begin{table}[!ht]
\centering
\caption{\textbf{Paired cross-track gains on 3D scenes and procedural graphics ($0$--$100$ scale).}
Differences compare \ourmethod{} against the native baseline of the identical backbone across paired prompts ($n$), reporting both composite Judge Score ($\Delta S_{\mathrm{all}}$) and uncapped visual score ($\Delta S_{\mathrm{vis}}$).}
\label{tab:harness-track-scores}
\footnotesize
\setlength{\tabcolsep}{3pt}

\begin{tabular*}{\linewidth}{@{\extracolsep{\fill}}l l r r c c r@{}}
\toprule
\textbf{Track} & \textbf{Backbone} & $n$ & $\Delta S_{\mathrm{all}}$ & \textbf{95\% CI} & \textbf{W/L/T} & $\Delta S_{\mathrm{vis}}$ \\
\midrule
Graphics & Flash & 11 & $+27.2$ & $[9.0,45.3]$ & 10/1/0 & $+19.1$ \\
Graphics & Sonnet & 11 & $-5.4$ & $[-20.8,10.1]$ & 6/5/0 & $-4.3$ \\
Graphics & Luna & 11 & $+6.8$ & $[-1.2,14.8]$ & 8/3/0 & $+6.5$ \\
Graphics & Astra & 11 & $+30.4$ & $[8.6,52.2]$ & 10/1/0 & $+28.2$ \\
Scenes & Flash & 10 & $+19.4$ & $[0.2,38.6]$ & 9/1/0 & $+15.4$ \\
Scenes & Sonnet & 10 & $+0.1$ & $[-31.1,31.3]$ & 3/6/1 & $+2.8$ \\
Scenes & Luna & 7 & $+14.6$ & $[2.5,26.8]$ & 7/0/0 & $+9.1$ \\
\bottomrule
\end{tabular*}
\par\raggedright
Flash, Sonnet, Luna, and Astra denote Gemini~3.8~Flash, Claude Sonnet~5, GPT-6~Luna, and GPT-6~Astra.
For graphics, $S_{\mathrm{vis}}$ retains all criterion weights prior to aggregate penalties and caps; for scenes, it additionally excludes the spatial-scale criterion.

\end{table}

\paragraph{Test-time verification compute and prompt-cache efficiency.}
Table~\ref{tab:harness-object-cost} reports execution rates and per-prompt token consumption across completed generation attempts (valuing CLI sessions at standard API list prices and excluding post-run evaluation calls; Claude Design is evaluated directly from its released reference programs).
Wrapping Gemini~3.8~Flash in \harness{} raises execution reliability from $93.3\%$ to \textbf{100.0\%}.
Crucially, because our modular workspace and stable tool schemas preserve prefix structure across agent turns, \textbf{$93.9\%$ of harness input tokens ($19{,}551.8$\,k of $20{,}816.0$\,k) are served from low-cost prompt cache}.
At an average cost of \$3.31 per object, test-time closed-loop verification enables a lightweight Flash backbone to achieve an $86.6$ Judge Score---substantially outperforming both single-pass Claude Opus~5.5 ($77.2$) and Claude Design ($37.4$).

\paragraph{Evaluator precision vs.\ generative exploration variance.}
In the auxiliary studies of Appendix~\ref{app:harness-extra}, each artifact is rebuilt, rendered, and scored under matched conditions with two independent judge samples.
Across repeated evaluations of fixed artifacts, the VLM judge exhibits a standard deviation of only $\sigma_{\mathrm{judge}} = 0.030$ for Gemini~3.1~Pro ($0.083$ for Gemini~3.8~Flash) on a $[0,1]$ scale, confirming that measurement noise is an order of magnitude smaller than the system-level gains reported in \S\ref{sec:experiments}.

\begin{table}[ht]
\centering
\caption{\textbf{Judge precision vs.\ structural exploration across repeated runs.}
Left: paired statistics across two independent runs under identical settings, juxtaposed with Gemini~3.1~Pro judge repeatability ($\sigma_{\mathrm{judge}} = 0.030$).
Right: an illustrative prompt in which two independent runs synthesize valid but topologically distinct decompositions (a single monolithic mesh vs.\ a $10$-part assembly).}
\label{tab:aa}
\small
\begin{minipage}[t]{0.48\textwidth}
\centering
\setlength{\tabcolsep}{5pt}
\begin{tabular}{@{}l r@{}}
\toprule
\textbf{Repeatability metric} & \textbf{Value} \\
\midrule
$\sigma_{\mathrm{judge}}$, Gemini~3.1~Pro ($3\times$) & \textbf{0.030} \\
$\sigma_{\mathrm{judge}}$, Gemini~3.8~Flash ($3\times$) & 0.083 \\
\midrule
Paired A/A runs $n$ & 6 \\
Mean difference $\bar{\Delta}$ & $+0.043$ \\
Median difference & $-0.008$ \\
Std.\ deviation $\sigma_{\Delta}$ & 0.202 \\
Std.\ error of $\bar{\Delta}$ & 0.082 \\
\bottomrule
\end{tabular}
\end{minipage}\hfill
\begin{minipage}[t]{0.48\textwidth}
\centering
\setlength{\tabcolsep}{7pt}
\begin{tabular}{@{}l r r@{}}
\toprule
\textbf{Divergent runs} & \textbf{Run A} & \textbf{Run B} \\
\midrule
Planned part count  & \textbf{1} & \textbf{10} \\
Refinement rounds   & 1 & 2 \\
Mesh triangles      & 4{,}796 & 61{,}340 \\
Agent tool calls    & 28 & 168 \\
Generation cost     & \$0.33 & \$1.61 \\
Stopped by & threshold & budget \\
\midrule
Fixed-judge score   & 0.906 & 0.700 \\
\bottomrule
\end{tabular}
\end{minipage}
\end{table}

Independent generation runs on open-ended prompts naturally explore diverse structural decompositions: in an A/A study across six paired runs with early pass stopping (Table~\ref{tab:aa}; $\sigma_{\Delta} = 0.202$), as well as a $12$-pair articulated study with a fixed plan ($\bar{\Delta} = -0.046$, $\sigma_{\Delta} = 0.225$, $2\,\text{SE} = \pm 0.130$), a single prompt may be realized either as a compact monolithic mesh ($1$ part, $4{,}796$ triangles) or as a fine-grained multi-part assembly ($10$ parts, $61{,}340$ triangles).
This structural diversity motivates two core design choices in our work:
(1)~within \harness{}, best-of-$N$ candidate initialization and multi-round refinement (\S\ref{app:harness-rounds}) hedge against single-rollout topological variance; and
(2)~in our experimental design, while full-workflow gains of $+0.15$ to $+0.49$ are cleanly resolved on $30$--$60$ paired prompts, fine-grained infrastructure optimizations are evaluated directly via high-power event-level telemetry across hundreds to thousands of invocations (\S\ref{app:loss-events}).

\subsection{Additional open-model results}
\label{app:model-baselines}

\paragraph{Executability and geometric fidelity.}
Table~\ref{tab:sft-full} reports per-dialect execution rates and geometric fidelity (F@0.05 and DINOv2-L similarity) across all fine-tuned checkpoints under the evaluation protocol of Table~\ref{tab:sft}.

\begin{table}[t]
\centering
\providecommand{\tbd}{\textcolor{black!35}{\ldots}}
\caption{\textbf{All fine-tuning mixtures.} Full results behind Table~\ref{tab:sft}, including every training mixture and per-dialect F@0.05.
All models are decoded at $T{=}0.7$ with one seed and scored on the same fixed test sets.
\emph{Exec}: programs that run in the dialect's runtime (\%); for GLSL, shaders that compile.
\emph{F}: F-score at $\tau{=}0.05$ between the generated and reference meshes, with failed programs scored 0.
\emph{DINO}: DINOv2-L cosine similarity between four-view renders of the generated and reference meshes (chance 0.159).
Best model result per column in bold.}
\label{tab:sft-full}
\scriptsize
\setlength{\tabcolsep}{3pt}
\resizebox{\textwidth}{!}{%
\begin{tabular}{@{}l l ccc cc cc cc c c@{}}
\toprule
& & \multicolumn{3}{c}{\bench{} (212)} & \multicolumn{2}{c}{Blender (103)} & \multicolumn{2}{c}{CadQuery (200)} & \multicolumn{2}{c}{OpenSCAD (50)} & GLSL (200) & Three.js (40) \\
\cmidrule(lr){3-5}\cmidrule(lr){6-7}\cmidrule(lr){8-9}\cmidrule(lr){10-11}\cmidrule(lr){12-12}\cmidrule(lr){13-13}
\textbf{Model} & \textbf{Training data} & Exec & F & DINO & Exec & F & Exec & F & Exec & F & Exec & Exec \\
\midrule
\multicolumn{13}{@{}l}{\textit{Zero-shot}} \\
Qwen3.5-9B & --- & 0.0 & .000 & .000 & 1.9 & .008 & 10.5 & .030 & 50.0 & .261 & 33.0 & 72.5 \\
Qwen3.8-27B & --- & 51.9 & .214 & .211 & 74.8 & .439 & 65.5 & .159 & 80.0 & \textbf{.415} & 57.0 & 72.5 \\
\midrule
\multicolumn{13}{@{}l}{\textit{Qwen3.5-9B + LoRA SFT}} \\
 & 5.2\,K, Blender & 73.1 & .257 & .228 & \textbf{93.2} & .694 & 31.5 & .081 & 32.0 & .099 & 23.0 & 47.5 \\
 & 102\,K, 4 dialects & 84.4 & .328 & .269 & \textbf{93.2} & .725 & 99.0 & .268 & 70.0 & .276 & 61.0 & 0.0 \\
 & 270\,K, 4 dialects & 84.9 & .317 & .266 & 90.3 & .693 & 98.5 & \textbf{.308} & 86.0 & .329 & 56.0 & 32.5 \\
\midrule
\multicolumn{13}{@{}l}{\textit{Qwen3.8-27B + LoRA SFT}} \\
 & 76\,K, CadQ./GLSL-heavy & 53.3 & .224 & .222 & 92.2 & .795 & \textbf{99.5} & .227 & 76.0 & .349 & 63.0 & 87.5 \\
 & 104\,K, Blender 40\% & \textbf{92.9} & .365 & \textbf{.303} & 91.3 & .786 & 99.0 & .226 & 72.0 & .298 & \textbf{73.0} & \textbf{92.5} \\
 & 104\,K + exec-DPO & \textbf{92.9} & \textbf{.374} & .298 & \textbf{93.2} & \textbf{.798} & 98.5 & .234 & \textbf{90.0} & .314 & 71.5 & \textbf{92.5} \\
\bottomrule
\end{tabular}}
\end{table}

Zero-shot Qwen3.5-9B executes \textbf{0/212} \bench{} tasks: $165$ of $212$ scripts fail immediately on deprecated scene-clearing operators, and patching that single call lifts execution only to $4/212$ because of wider Blender~5.0 API drift.
After fine-tuning, 9B execution on the held-out Blender suite reaches \textbf{93.2\%}, matching the $96/103$ pass rate of the reference scripts themselves, and F@0.05 rises from $0.000$ to $0.328$ on \bench{} and from $0.008$ to $0.725$ on Blender.
On Qwen3.8-27B, fine-tuning nearly doubles unconditional F@0.05, from $0.214$ to \textbf{0.374} on \bench{} ($+74.8\%$ relative) and from $0.439$ to \textbf{0.798} on Blender ($+81.8\%$); whereas the zero-shot checkpoint succeeds only on the easiest half of prompts (conditional F@0.05 of $0.420$), the fine-tuned model executes an additional $41.0\%$ of tasks while keeping conditional F@0.05 at $0.397$--$0.424$.
The $104$\,K mixture, which adds execution-verified bootstrap programs and balances dialects, further lifts GLSL compilation to \textbf{73.0\%} ($+16.0$\,pp) and Three.js execution to \textbf{92.5\%} ($+20.0$\,pp).

\paragraph{Post-training with execution, geometry, and termination DPO.}
\label{app:preference}
We construct preference pairs by sampling multiple completions per training prompt and ranking them along three verifiable axes---\emph{execution success} (running vs.\ failing scripts), \emph{geometric fidelity} (highest vs.\ lowest F@0.05 with $\Delta\text{F} \ge 0.15$), and \emph{sequence termination}---and optimize the merged SFT checkpoints with DPO~\citep{rafailov2023dpo} (Tables~\ref{tab:dpo-full} and~\ref{tab:bestofn}; mixture specifications in Table~\ref{tab:mixes}).

\begin{table}[ht]
\centering
\caption{\textbf{Direct preference optimization across verifiable feedback signals (greedy decoding).}
\emph{Execution feedback} pairs an executing completion against a failing completion for the same prompt; \emph{geometry feedback} pairs the highest- and lowest-F@0.05 completions; and \emph{termination feedback} pairs a non-terminating greedy rollout against the shortest valid completion.
All runs in this table use greedy decoding ($T{=}0$); multi-seed sampling ($T{=}0.7$, pass@4) is reported in Table~\ref{tab:bestofn}.}
\label{tab:dpo-full}
\small
\setlength{\tabcolsep}{4.5pt}
\begin{tabular}{@{}l r rr rr rr@{}}
\toprule
& & \multicolumn{2}{c}{\bench{} exec} & \multicolumn{2}{c}{\bench{} geometry} & \multicolumn{2}{c}{Blender held-out} \\
\cmidrule(lr){3-4}\cmidrule(lr){5-6}\cmidrule(lr){7-8}
\textbf{Base $\rightarrow$ +DPO} & \textbf{pairs} & before & after & F$^{\text{ok}}$ & CD\,$\downarrow$ & F$^{\text{ok}}$ & CD\,$\downarrow$ \\
\midrule
\multicolumn{8}{@{}l}{\textit{execution feedback}} \\
9B, 5.2\,K base $\rightarrow$ round 1 & 2{,}590 & 75.0--82.1 & \textbf{93.9} & 0.384 & 0.201 & --- & --- \\
\hspace{1em}$\rightarrow$ round 2     &   587 & 93.9 & \textbf{96.2} & 0.383 & 0.200 & --- & --- \\
9B, 102\,K base                       & 1{,}052 & 84.9 & 87.3 & 0.375 & 0.202 & 0.783 & 0.077 \\
9B, 270\,K base                       &   909 & 90.1 & 89.6 & 0.388 & 0.204 & 0.778 & 0.078 \\
27B, 104\,K base                      &   756 & \textbf{93.4} & 91.5 & 0.395 & 0.195 & 0.872 & 0.052 \\
\midrule
\multicolumn{8}{@{}l}{\textit{geometry feedback}} \\
9B, 270\,K base                       & 2{,}357 & 90.1 & 88.2 & \textbf{0.403} & \textbf{0.195} & \textbf{0.800} & \textbf{0.070} \\
\midrule
\multicolumn{8}{@{}l}{\textit{termination feedback}} \\
9B, 310\,K base $\rightarrow$ round 1 & 1{,}779 & 79.2 & 85.9 & 0.373 & 0.206 & 0.885 & 0.054 \\
\hspace{1em}$\rightarrow$ round 2     &   478 & 85.9 & 84.9 & 0.369 & 0.205 & 0.876 & 0.055 \\
\bottomrule
\end{tabular}
\end{table}

Each preference signal targets a distinct aspect of code generation quality:
(1)~\emph{Execution-feedback DPO} substantially increases single-sample reliability and multi-sample coverage: on the 9B ($5.2$\,K) checkpoint, two rounds of execution DPO lift greedy \bench{} execution from $75.0\%$ to \textbf{96.2\%} and raise four-seed union pass@4 ($T{=}0.7$) from $91.0\%$ to \textbf{99.1\%} (Table~\ref{tab:bestofn}).
On the flagship 27B ($104$\,K) checkpoint, execution DPO preserves top \bench{} execution (\textbf{92.9\%} at $T{=}0.7$) while achieving the highest overall geometric fidelity on both \bench{} (F@0.05 of \textbf{0.374}) and Blender (\textbf{93.2\%} execution, \textbf{0.798} F@0.05) and lifting OpenSCAD execution by $+18.0$\,pp (from $72.0\%$ to \textbf{90.0\%}; Table~\ref{tab:sft-full}).
(2)~\emph{Geometry-feedback DPO} directly refines 3D shape fidelity once execution is saturated: on the 9B ($270$\,K) checkpoint, it lifts conditional F@0.05 from $0.387$ to \textbf{0.403} while achieving a $98.1\%$ pass@4 execution rate (Table~\ref{tab:bestofn}).
(3)~\emph{Termination-feedback DPO} eliminates repetitive brace-closing loops under greedy decoding.
Pairing non-terminating greedy outputs with the model's shortest valid completion for the same training prompt across two lightweight DPO rounds ($1{,}779$ and $478$ pairs) compresses mean OpenSCAD output length from $30{,}163$ to $2{,}884$ tokens and drives greedy execution from $8.0\%$ to \textbf{86.0\%} ($+78.0$\,pp) on OpenSCAD, from $59.0\%$ to \textbf{89.5\%} ($+30.5$\,pp) on GLSL, and from $79.2\%$ to \textbf{84.9\%} on \bench{}.

\begin{table}[ht]
\centering
\caption{\textbf{Best-of-$N$ sampling} at $T{=}0.7$, four seeds.
Execution-feedback DPO lifts both the mean seed and the union: pass@4 reaches 99.1\%.}
\label{tab:bestofn}
{\small
\setlength{\tabcolsep}{6pt}
\begin{tabular}{@{}l rrrr r@{}}
\toprule
\textbf{best-of-$N$ ($T{=}0.7$, $N{=}4$)} & \multicolumn{4}{c}{per-seed exec} & pass@4 \\
\midrule
9B, 5.2\,K base            & 76.4 & 66.5 & 64.6 & 75.0 & 91.0 \\
\quad + exec-DPO           & 90.6 & 93.4 & 86.3 & 88.2 & \textbf{99.1} \\
9B, 270\,K + geo-DPO       & 88.2 & 83.5 & 82.1 & 86.8 & 98.1 \\
\bottomrule
\end{tabular}}
\end{table}

\paragraph{Multi-dialect synergy and mixture balancing.}
Comparing single-dialect and multi-dialect training mixtures reveals strong cross-language transfer alongside clear mixture-design principles:
(1)~\emph{Cross-dialect transfer}: joint multi-dialect training boosts CadQuery execution to \textbf{98.0\%}, compared with $73.0\%$ under single-dialect training.
(2)~\emph{Target-dialect upsampling}: increasing the Blender share of the 9B mixture from $15\%$ to $34\%$ (Table~\ref{tab:mixes}, $116$\,K balanced) raises held-out Blender F@0.05 from $0.705$ to \textbf{0.816} (with failed builds scored zero) and lifts greedy OpenSCAD execution from $4\%$ to $56\%$.
(3)~\emph{Capacity and mixture balance for long-horizon scripts}: while isolated fine-tuning of a 9B checkpoint on $280$ long multi-part Three.js programs approaches the sequence window and induces non-termination, combining balanced multi-dialect supervision with larger capacity in the 27B ($104$\,K) mixture elevates Three.js execution from $72.5\%$ to \textbf{92.5\%} (Table~\ref{tab:sft-full}).

\subsection{Additional harness results}

\paragraph{Numerical results for Figure~\ref{fig:main-results}.}
Table~\ref{tab:harness} reports the exact numerical values underlying Figure~\ref{fig:main-results}a--c alongside the VLM Preference and SigLIP-2 text--image similarity metrics.

\begin{table}[t]
\centering
\caption{
\textbf{3D object generation.}
Numerical values of Figure~\ref{fig:main-results}a--c.
Comparisons at a fixed backbone (upper) and against Claude Design (lower), on identical prompts.
SigLIP-2 reports cosine similarity; Judge Score reports overall rubric quality (0--100).
}
\label{tab:harness}
\footnotesize
\setlength{\tabcolsep}{3pt}

\begin{tabular*}{\linewidth}{@{\extracolsep{\fill}}l l c c c r@{}}
\toprule
\textbf{Model} & \textbf{Harness} & \textbf{Human Pref. (\%)}$\uparrow$ & \textbf{VLM (\%)}$\uparrow$ & \textbf{SigLIP-2}$\uparrow$ & \textbf{Judge Score}$\uparrow$ \\
\midrule
\multicolumn{6}{@{}l}{\textbf{Same Model with Different Harnesses}} \\
Gemini~3.8~Flash & Gemini CLI & 13\% & 15\% & 0.1356 & 61.3 \\
Gemini~3.8~Flash & \ourmethod{} & 87\% & 85\% & 0.1468 & 86.6 \\
Claude Opus~5.5 & Claude Code & 21\% & 19\% & 0.1437 & 77.2 \\
Claude Opus~5.5 & \ourmethod{} & 79\% & 81\% & 0.1490 & 88.5 \\
GPT-6 Astra & Codex & 31\% & 28\% & 0.1456 & 76.4 \\
GPT-6 Astra & \ourmethod{} & 69\% & 72\% & 0.1488 & 87.7 \\
GPT-6 Luna & Codex & 11\% & 16\% & 0.1293 & 64.4 \\
GPT-6 Luna & \ourmethod{} & 89\% & 84\% & 0.1341 & 85.9 \\
\midrule
\multicolumn{6}{@{}l}{\textbf{System comparison: \ourmethod{} vs.\ Claude Design}} \\
Claude Opus~5.5 & Claude Design & 9\% & 13\% & 0.1438 & 37.4 \\
Gemini~3.8~Flash & \ourmethod{} & 91\% & 87\% & 0.1468 & 86.6 \\
\hline
Claude Opus~5.5 & Claude Design & 10\% & 14\% & 0.1438 & 37.4 \\
Claude Opus~5.5 & \ourmethod{} & 90\% & 86\% & 0.1493 & 87.7 \\
\bottomrule
\end{tabular*}

\end{table}

\label{app:harness-extra}

\begin{table}[!ht]
\centering
\caption{\textbf{\harness{} vs.\ one-shot generation across models.}
Scores are assigned by a fixed judge with two samples per output.
In both settings, the GPT-6 models use the Codex CLI backend and Gemini~3.8~Flash uses the Gemini CLI backend.
$\Delta$ is the paired score difference in points, and W/L/T counts per-prompt wins, losses, and ties by score difference.
The generation settings differ from those of the object evaluation in Figure~\ref{fig:main-results}a.}
\label{tab:harness-development}
\small
\setlength{\tabcolsep}{3.5pt}

\begin{tabular*}{\linewidth}{@{\extracolsep{\fill}}l l r r c c@{}}
\toprule
\textbf{Model} & \textbf{Prompts} & $n$ & $\Delta$ & \textbf{95\% CI} & \textbf{W/L/T} \\
\midrule
GPT-6 Luna & hard & 10 & $+29.1$ & $[14.1,44.1]$ & 10/0/0 \\
GPT-6 Terra & hard & 11 & $+19.1$ & $[1.3,36.8]$ & 8/2/1 \\
GPT-6 Sol & hard & 11 & $+15.5$ & $[3.1,27.9]$ & 8/2/1 \\
Gemini~3.8~Flash & hard & 8 & $+40.1$ & $[20.9,59.2]$ & 7/0/1 \\
Gemini~3.8~Flash & mixed & 8 & $+52.3$ & $[34.5,70.1]$ & 8/0/0 \\
\bottomrule
\end{tabular*}

\end{table}

\paragraph{Multi-round refinement and backbone scaling across tracks.}
Tables~\ref{tab:harness-development} and~\ref{tab:harness-arms} evaluate \harness{} across five model backbones on challenging object, articulated-mechanism, and shader prompts.
Across all five configurations in Table~\ref{tab:harness-development}, \harness{} delivers consistent paired gains of $+15.5$ to $+52.3$ points over one-shot generation, achieving \textbf{100\% execution reliability} across every model in Table~\ref{tab:harness-arms} (compared with $38\%$--$92\%$ for one-shot baselines).
Within multi-round trajectories, closed-loop refinement boosts mean articulated-object scores from $0.605$ (Round~$0$) to $0.789$ (final selected round)---gaining $+0.284$ on the hardest mechanisms with $58\%$ achieving a full rubric pass---and raises GLSL shader scores from $0.712$ to $0.748$ ($5/6$ passing).

Crucially, the paired comparison in the lower panel of Table~\ref{tab:harness-arms} demonstrates how harness verification interacts with backbone capability on hard prompts:
(1)~\emph{Bridging model tiers}: equipping the lightweight Gemini~3.8~Flash model with \harness{} raises its mean score from $0.146$ to $0.625$, matching the frontier GPT-6~Sol one-shot baseline ($0.603$) and winning $5$ of $8$ head-to-head prompts.
(2)~\emph{Scaling with frontier backbones}: applying \harness{} to GPT-6~Sol itself pushes performance further to \textbf{0.758} ($+15.5$ points over Sol one-shot, winning $8/8$ prompts against Flash + \harness{}), showing that closed-loop spatial verification both elevates lightweight models to frontier one-shot quality and unlocks substantially higher ceilings on flagship models.

\begin{table}[ht]
\centering
\caption{\textbf{Mean score, execution reliability, and cost per run} on the hard-prompt suite, comparing \harness{} against one-shot generation.
Execution rate is the percentage of programs that build and export valid geometry.
Lower panel: paired comparisons on eight matched hard prompts under the same fixed judge (score differences in points on a $0$--$100$ scale), illustrating both tier-bridging (Flash + \harness{} vs.\ GPT-6~Sol one-shot) and frontier scaling (GPT-6~Sol + \harness{}).}
\label{tab:harness-arms}
\vspace{-1mm}
{\small
\setlength{\tabcolsep}{5pt}
\begin{tabular}{@{}l cc cc cc@{}}
\toprule
& \multicolumn{2}{c}{Score} & \multicolumn{2}{c}{Execution} & \multicolumn{2}{c}{\$ / run} \\
\cmidrule(lr){2-3}\cmidrule(lr){4-5}\cmidrule(lr){6-7}
\textbf{Model} & Harness & 1-shot & Harness & 1-shot & Harness & 1-shot \\
\midrule
GPT-6 Sol   & \textbf{0.758} & 0.603 & \textbf{100\%} & 100\% & 6.72 & 0.29 \\
GPT-6 Terra & \textbf{0.646} & 0.441 & \textbf{100\%} & 92\% & 3.99 & 0.05 \\
GPT-6 Luna  & \textbf{0.601} & 0.289 & \textbf{100\%} & 75\% & 1.05 & 0.01 \\
Gemini 3.8 Flash & \textbf{0.625} & 0.146 & \textbf{100\%} & 38--57\% & 2.41 & 0.04 \\
Claude (Claude Code default, 1-shot) & --- & 0.472 & --- & 75\% & --- & 1.17 \\
\bottomrule
\end{tabular}}

\vspace{2.5mm}
{\small\textbf{Bridging model tiers and scaling with frontier backbones} (paired on $n=8$ hard prompts):\par}
\vspace{1mm}
{\small
\setlength{\tabcolsep}{5pt}
\begin{tabular}{@{}l r r c c@{}}
\toprule
\textbf{Paired comparison} & $n$ & $\Delta$ \textbf{(pts)} & \textbf{95\% CI} & \textbf{W/L/T} \\
\midrule
Gemini~3.8~Flash (harness) vs.\ GPT-6~Sol (1-shot) & 8 & $-0.2$  & $[-11.8, +11.4]$ & \textbf{5/3/0} \\
GPT-6~Sol (harness) vs.\ Gemini~3.8~Flash (harness) & 8 & $+16.9$ & $[+10.8, +22.9]$ & \textbf{8/0/0} \\
\bottomrule
\end{tabular}}
\end{table}

\paragraph{Explicit planning of kinematic joint couplings.}
\label{app:coupling-study}
Complex mechanisms such as scissor lifts, parallel grippers, and synchronized double doors require coordinated multi-joint motion via URDF \texttt{mimic} constraints ($q_j = m_j q_i + b_j$).
On ten coupled-linkage prompts evaluated at Round~$0$ ($T=0$), augmenting the planner schema with an explicit coupling field increases valid URDF \texttt{mimic} declarations from \textbf{3/10 (30\%) to 10/10 (100\%)}: eight mechanisms correctly reduce to a single independent degree of freedom, and the remaining two appropriately retain secondary degrees of freedom for wheel rotation or mirror tilt.
Because Round~$0$ precedes iterative joint-sweep refinement (where both arms begin at comparable initial visual scores, $\Delta = -0.006$), this ablation highlights a clean architectural division of labour: explicit schema planning guarantees topological and kinematic coupling completeness in Round~$0$, establishing a valid mechanism graph for subsequent joint-sweep collision refinement (\S\ref{app:harness-design}).

\begin{table}[!ht]
\centering
\caption{\textbf{Component-level systems ablations and large-scale telemetry analysis.}
Each infrastructure mechanism in \harness{} targets a specific reliability, topological, or compute-efficiency bottleneck and is evaluated on its direct operational metric across large-scale invocation logs.}
\label{tab:loss-events}
\small
\setlength{\tabcolsep}{4pt}
\begin{tabular}{@{}l l r r r l@{}}
\toprule
\textbf{Mechanism} & \textbf{Metric} & $N$ & \textbf{Without} & \textbf{With} & \textbf{Effect} \\
\midrule
Plan resampling on reset & Invalid plans & 560 calls & 4.7\% & \textbf{0.7\%} & $-85\%$ ($p{=}0.0067$) \\
\quad dangling-link trigger & Dangling refs. & 416 calls & 2.1\% & \textbf{0.0\%} & $-100\%$ ($p{=}0.038$) \\
Coupling schema (\S\ref{app:coupling-study}) & Valid URDF \texttt{mimic} & 10 tasks & 3/10 & \textbf{10/10} & $+70$\,pp \\
Verdict/transport split & Redundant retries & 3{,}677 calls & 26.6\% & \textbf{1.3\%} & $-95\%$ \\
\quad + bounded payloads & Uncached tok./req. & 5{,}194 reqs & 38{,}521 & \textbf{13{,}480} & $-65\%$ \\
\quad + bounded payloads & Median cost/round & 142 rounds & \$1.572 & \textbf{\$0.950} & $-40\%$ \\
Scoped session horizon & Session tok./cost & 1{,}063 sess. & full & \textbf{$0.5\times$ turns} & $-68\%$ / $-55\%$ \\
\bottomrule
\end{tabular}

\vspace{1.2em}
{\captionsetup{type=table}\centering
\caption{\textbf{Preservation of articulated door-and-handle geometry across $12$ robot manipulation cases} (two assets $\times$ two initial states $\times$ three control regimes).
Code simulation and its reconstruction through UWM's frozen VAE both preserve assessable rigid contours in $12/12$ cases, whereas $29$ of $36$ UWM rollouts suffer contour degradation.}
\label{tab:code-vs-wam}
\small
\begin{tabular}{@{}l r r@{}}
\toprule
\textbf{Rollout} & \textbf{Criterion met} & \textbf{Degraded / abstained} \\
\midrule
3D code simulated in MuJoCo & \textbf{12 / 12 (100\%)} & 0 \\
Code renders through UWM's frozen VAE & \textbf{12 / 12 (100\%)} & 0 \\
Static last frame (no motion) & 12 / 12 (100\%) & 0 \\
UWM action-conditioned rollout ($3$ seeds per case) & 7 / 36 (19.4\%) & 29 \\
\bottomrule
\end{tabular}
}
\end{table}

\paragraph{Targeted ablations of harness infrastructure.}
\label{app:loss-events}
Whereas end-to-end scores resolve workflow gains on $30$--$60$ prompts, individual infrastructure mechanisms are evaluated with event-level telemetry over hundreds to thousands of invocations (Table~\ref{tab:loss-events}).
(1)~\emph{Context-reset plan resampling}: when a planner emits a degenerate kinematic hierarchy whose joints reference undeclared links, resetting the context to the original request with the violated schema rule, instead of repairing in-context, cuts the plan-stage failure rate by \textbf{6.7$\times$} ($13/277$ to $2/276$; two-sided Fisher's exact test, $p=0.0067$) at negligible per-call cost (\$0.0337 vs.\ \$0.0345), and a reset triggered specifically on dangling link references eliminates that failure mode ($0/276$ vs.\ $3/140$, $p=0.038$).
(2)~\emph{Verdict--transport decoupling}: reporting a valid negative observation, such as a detected collision, as an MCP transport error triggers blind retry loops and, in some agent backends, base64-serialized images; separating semantic verdicts from protocol errors and bounding diagnostic payloads (\S\ref{app:harness-tools}) reduces redundant retries by \textbf{20$\times$} across $3{,}677$ invocations on $14$ matched prompts, uncached prompt tokens per request by \textbf{65\%}, and median generator cost per round by \textbf{40\%}, while preserving artifact quality ($\Delta = +0.015$).
(3)~\emph{Scoped session horizons}: across $1{,}063$ sessions ($30{,}401$ turns), per-turn input grows linearly with depth as $\operatorname{input}(t) \approx 12{,}364 + 1{,}822\,t$, so cumulative session tokens scale as $n^{1.60}$ over $n$ turns; halving turn depth cuts cumulative tokens by \textbf{68\%} and net API cost by \textbf{55\%} after prompt-cache discounts, which motivates the file-disjoint subtree decomposition of \S\ref{app:harness-plan}.

\FloatBarrier

\FloatBarrier
\section{Executable articulation and comparison with UWM}
\label{app:code-vs-wam}

\paragraph{Explicit kinematic simulation vs.\ latent video rollouts.}
Complementing the robotics application in \S\ref{sec:applications}, we examine how executable 3D code preserves rigid-link structure during robot manipulation compared with a frozen video world-action model, UWM~\citep{zhu2025uwm}.
Using an articulated microwave asset synthesized by GPT-6~Astra and verified by \harness{} tooling (passing forward-kinematics, rigid-distance, and hinge-sweep gates across five poses), we instantiate a MuJoCo simulation alongside the native reference asset across two initial contact states and three robot control sequences---\emph{contact push}, \emph{hold}, and \emph{retreat}---yielding $12$ evaluation cases.
Each rollout executes $16$ matched robot action commands at $20$\,Hz: UWM generates three seeded video rollouts per case ($36$ total) from observation history and action commands, whereas code simulation steps the explicit articulated asset directly inside MuJoCo.

\paragraph{Quantitative evaluation of structural preservation.}
To measure whether rollouts maintain coherent rigid-part geometry under physical interaction, we apply a calibrated orientation-aware door-and-handle contour detector across all outputs.
Calibrated prior to evaluation on $24$ held-out static renders across six hinge angles and four robot poses, the detector recovers the true hinge angle within $1^\circ$ on all $20$ accepted detections and abstains whenever contours degrade or become ambiguous.
To isolate latent dynamics drift from autoencoder reconstruction blur, we additionally evaluate a \emph{VAE-reconstructed code control} by passing every simulator render through UWM's frozen VAE encoder and decoder.
Across all $72$ evaluated outputs (Table~\ref{tab:code-vs-wam}), both direct code simulation and VAE-reconstructed code renders satisfy the geometric criterion in \textbf{12/12 (100\%)} rollouts, whereas UWM predictions satisfy the criterion in only \textbf{7/36 (19.4\%)} rollouts ($29/36$ fail to preserve assessable door-and-handle contours).
Because passing code renders through UWM's own frozen VAE preserves a $12/12$ pass rate, the contour degradation in UWM rollouts stems from latent action dynamics rather than VAE reconstruction limits.

\begin{figure}[!tbp]
\centering
\includegraphics[width=\textwidth]{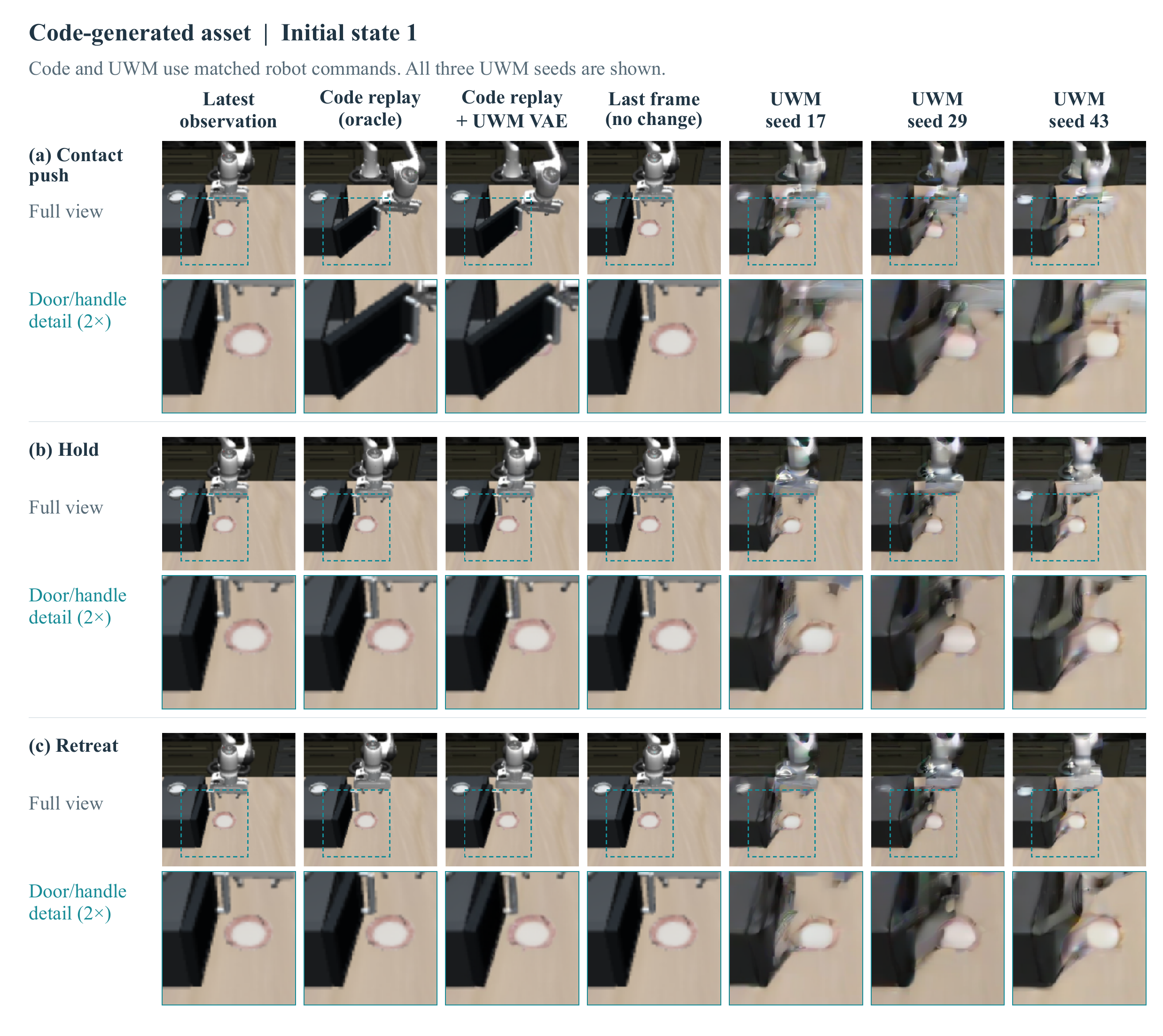}
\caption{\textbf{Articulated code simulation vs.\ UWM rollouts on the generated asset (Initial State 1).}
Rows correspond to contact push, hold, and retreat control sequences ($16$ commands at $20$\,Hz).
Columns show the initial observation, MuJoCo simulation of the generated 3D code, its reconstruction through UWM's frozen VAE, the static last-frame reference, and all three UWM rollout seeds.
Enlarged $2\times$ crops (dashed boxes) demonstrate that executable 3D code preserves sharp, consistent door and handle geometry both before and after VAE reconstruction, whereas UWM rollouts exhibit seed-dependent contour drift.}
\label{fig:code-wam-generated-state1}
\end{figure}

\begin{figure}[p]
\centering
\includegraphics[width=\textwidth]{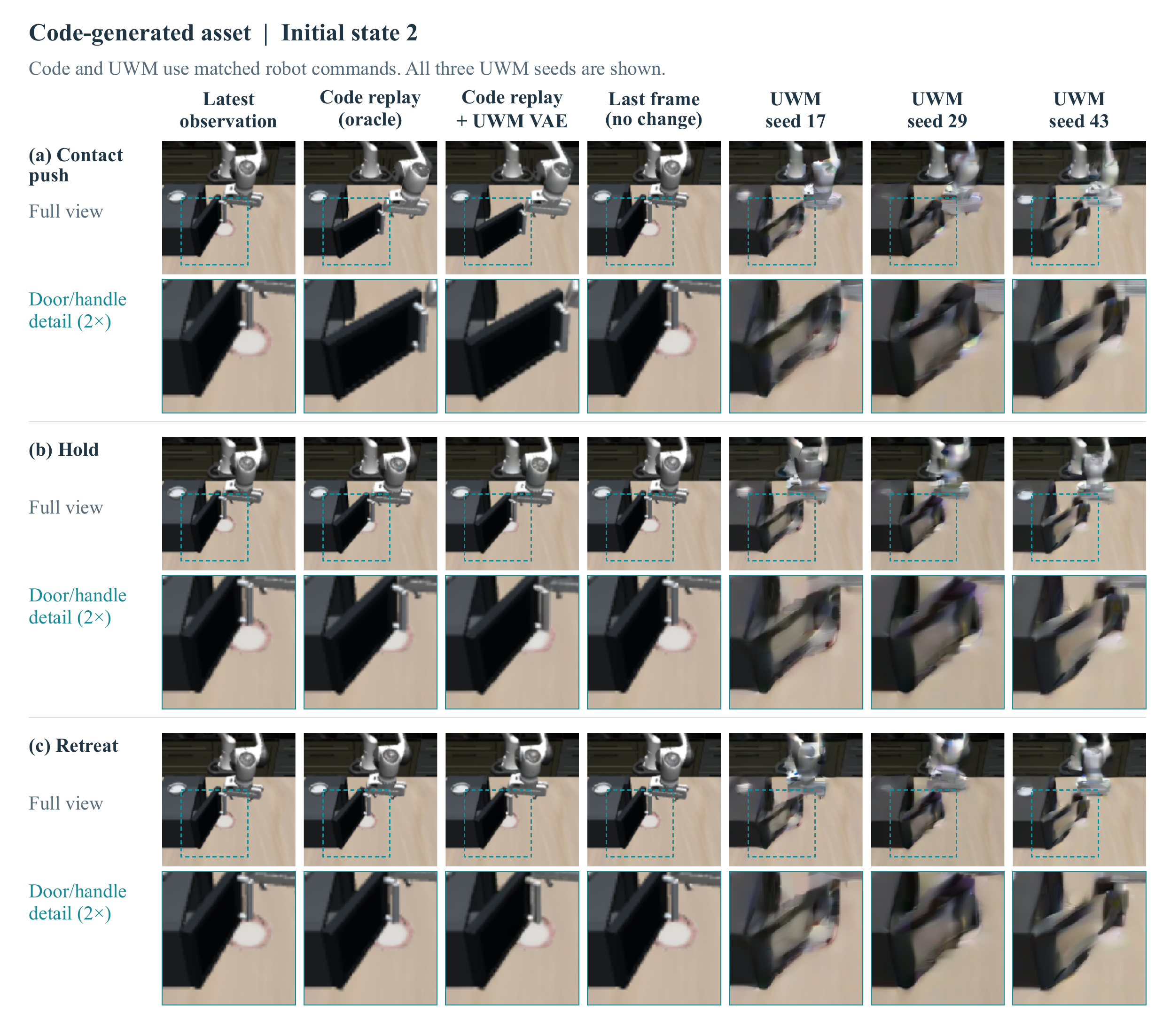}
\caption{\textbf{Articulated code simulation vs.\ UWM rollouts on the generated asset (Initial State 2).}
Control sequences, column layout, and fixed $2\times$ crops match Figure~\ref{fig:code-wam-generated-state1}, displaying all nine UWM rollouts for Initial State~2.}
\label{fig:code-wam-generated-state2}
\end{figure}

\paragraph{Complementary strengths of simulator-ready 3D code.}
Table~\ref{tab:code-vs-wam} and Figures~\ref{fig:code-wam-generated-state1}--\ref{fig:code-wam-generated-state2} clarify the structural advantage of programmatic 3D assets for embodied simulation.
While a static last-frame baseline trivially preserves initial contours ($12/12$) by predicting zero motion, pixel-space world models synthesize motion directly in latent space at the cost of frequent contour distortion across random seeds.
By contrast, compiling generated 3D code into a physics simulator combines both properties: the simulator enforces exact rigid-body transformations, joint axes, and articulated state transitions over time while maintaining crisp, view-consistent geometry across contact push, hold, and retreat controls.

\end{document}